\pdfoutput=1
\def\ARXIV{}
\documentclass[11pt]{article}
\ifdefined\ARXIV
  \usepackage[preprint]{acl}
\else
  \usepackage[review]{acl}
\fi
\usepackage{times}
\usepackage{latexsym}
\usepackage[T1]{fontenc}
\usepackage[utf8]{inputenc}
\usepackage{microtype}
\usepackage{inconsolata}
\usepackage{graphicx}
\usepackage{booktabs}
\usepackage{amsmath}
\usepackage{amsfonts}
\usepackage{xcolor}
\usepackage{multirow}
\usepackage{makecell}
\usepackage{array}
\newcolumntype{L}[1]{>{\raggedright\arraybackslash}p{#1}}
\usepackage{stfloats}
\usepackage{listings}
\usepackage{needspace}
\usepackage{tikz}
\usetikzlibrary{arrows.meta,calc}

\newcommand{\method}{\textsc{HindAudit}}
\newcommand{\splus}{$S^{+}$}
\newcommand{\sminus}{$S^{-}$}
\newcommand{\szero}{$S^{0}$}
\newcommand{\ptrue}{$p_{\text{true}}$}
\newcommand{\chn}[1]{\tikz[baseline=(c.base)]\node[circle,fill=black,text=white,inner sep=0pt,minimum size=3.3mm,font=\scriptsize\bfseries](c){#1};}

\title{Where Does Hindsight Enter Clinical Agent Benchmarks?\\
An Audit of Simulators and Case Construction}

\ifdefined\ARXIV
\author{Meilu Zhu \\
  Department of Electrical and \\ Computer Engineering \\
  The University of Hong Kong \\
  \texttt{meiluzhu@hku.hk} \\\And
  Xiaohan Li \\
  Department of Restorative \\ Dental Sciences \\
  Faculty of Dentistry \\
  The University of Hong Kong \\
  \texttt{li2xh@hku.hk} \\\And
  Axiu Mao\thanks{Corresponding author.} \\
  Department of Infectious \\ Diseases and Public Health \\
  City University of Hong Kong \\
  \texttt{max.mao@my.cityu.edu.hk} \\}
\else
\author{Anonymous ACL submission}
\fi

\begin{document}
\maketitle
\begin{abstract}
Benchmarks for LLM diagnostic agents are built from retrospective records: the case an agent
reads is rewritten from discharge summaries, and unrecorded test results are generated by a
simulator that often sees the final diagnosis. \method{} separates three hindsight channels
(the diagnosis given to the simulator, the retrospective context it conditions on, and the
agent-facing case text) and measures each with judges trained only on real data. On 27{,}489
emergency admissions from MIMIC-IV, five registered tests found no leakage into simulated
laboratory results: providing the diagnosis to five simulators had a negligible effect, and
prompted gatekeepers diluted the diagnostic signal, although exploratory results suggest they
can raise agent scores. The largest effect lay in the case text. Against a presentation-time
reconstruction from triage records, benchmark-style summaries raised the
zero-shot accuracy of four LLMs from 25.4\% to 39.7\% and discharge histories to 45.3\%, and
raised six interactive agents' scores while barely changing their ranking. The exploratory gap
was reproduced in a registered replication on held-out patients. Masking diagnosis names left
it unchanged. Because MIMIC lacks a contemporaneous bedside history, the gap is an upper bound
on hindsight. We report all registered outcomes and recommend construction practices.
\end{abstract}

\section{Introduction}

Large language model (LLM) agents are increasingly evaluated as sequential diagnosticians:
the agent reads a short presentation of a patient, orders examinations, receives their
results, and commits to a diagnosis~\cite{schmidgall2024agentclinic,nori2025sdbench,qiu2025diaggym,hager2024mimiccdm}.
Scores on these benchmarks are used to compare models and to argue that agents are
approaching clinical usefulness, so whether a benchmark measures diagnosis under realistic
uncertainty is not a technicality: if the answer is easier to see than it would be at the
bedside, every score built on it is inflated.

No real patient can be re-examined, so these benchmarks are built from retrospective
electronic health records, and two construction steps are shared by most of them. The case
the agent reads is extracted or rewritten, often by an LLM, from notes written \emph{after}
the diagnosis was known, most commonly the discharge summary. Moreover, examinations the agent
requests but that were never performed are answered by a simulator (a fine-tuned ``world
model''~\cite{qiu2025diaggym} or a prompted ``gatekeeper'' LLM~\cite{nori2025sdbench}), which is
usually given the full record, including the final diagnosis. Both steps invite
\emph{hindsight}: information that exists only because the answer is known can reach what the
agent sees. Benchmark builders have focused on the second step and on its most visible form,
the diagnosis name. Simulators are instructed not to reveal the answer, and case texts that
name the diagnosis are filtered out~\cite{qiu2025diaggym,nori2025sdbench,hager2024mimiccdm}.
Whether they address the main source of hindsight is unknown.

We ask where hindsight actually enters, and how much. To answer this, we propose \method{}, an
audit that separates three channels (the diagnosis line in the simulator's input, the retrospective
context the simulator conditions on, and the case text the agent reads) and measures each against
real data: judges
trained only on real records score how strongly simulated results point to the true diagnosis,
and LLMs and text classifiers are compared across versions of the same case, including a
\emph{presentation-time reconstruction} built from the emergency-department triage record. The
tests of the simulator-side concern were registered before each run, and we report all of them.

On 27{,}489 emergency admissions from MIMIC-IV, the channel that builders guard most carefully
mattered least: providing the diagnosis to five simulators (seven configurations) had a
negligible effect (at most 1.4 pp) on the laboratory-result signal detected by real-data judges, and prompted gatekeepers diluted
that signal, although an exploratory analysis suggests that they can still raise the scores of
agents that rely on laboratory results. In the case text, four local LLMs diagnosed 25.4\% of presentation-time cases correctly, but 39.7\%
of the corresponding benchmark-style summaries and 45.3\% of discharge histories, a gap that
replicated on held-out patients under a registered rule and was unchanged when diagnosis names
were removed, and six interactive agents showed it too. As MIMIC lacks a contemporaneous bedside history, this gap is
an upper bound on hindsight. Whatever its source,
diagnosis-name filters cannot remove it. These results locate the main threat to validity
in case construction rather than simulators, and suggest building agent-facing cases from
presentation-time information. Our contributions are:
\begin{itemize}
  \item \textbf{An audit method and a reference construction.} \method{} measures three
  hindsight channels separately with real-data judges, and a
  presentation-time reconstruction from triage records provides a reference input.
  \item \textbf{The case text as the main channel.} Benchmark texts make diagnosis 12--22
  percentage points easier for each of five LLMs from three families. The gap survives
  diagnosis-name masking and replicates on held-out patients.
  \item \textbf{The simulator side, with registered tests.} Five registered tests found no
  leakage into simulated laboratory results: the diagnosis line had a negligible effect, and
  gatekeepers kept 16--62\% of the real signal (world models 70--84\%).
  \item \textbf{Two pitfalls for leakage audits.} Placeholder masking creates a cue for
  text-based judges, and comparing raw probabilities across classifiers can hide a large effect.
\end{itemize}

\section{Related work}

\paragraph{Simulated environments for diagnostic agents.}
AgentClinic~\cite{schmidgall2024agentclinic} uses LLM-played patients that know the
symptoms but not the diagnosis. MIMIC-CDM~\cite{hager2024mimiccdm} replays recorded results
only and masks diagnosis names in them. SDBench~\cite{nori2025sdbench} implements a
gatekeeper (o4-mini) ``with access to the full \ldots{} case file, including the final
diagnosis'', which synthesizes case-consistent findings for tests not in the case and is
described as removing hindsight bias from the case write-ups. DiagGym~\cite{qiu2025diaggym}
trains a world model on MIMIC-IV whose input profile, taken from the discharge summary,
includes ``the final diagnosis''. The agent instead reads an initial inquiry produced by an
LLM from the restructured record and the discharge note, filtered by asking whether it
``directly contain[s] the name'' of the diagnosis. EHR world models for long-horizon
trajectories~\cite{ehrworld2026} widen the use of learned simulators.

\paragraph{Leakage and evaluation validity.}
PULSE-CARE~\cite{pulsecare2026} showed that agent scores change substantially when the
patient simulator is swapped, so scores hold only conditional on the simulator. Explicit leakage is commonly addressed by filtering
diagnosis names~\cite{hager2024mimiccdm,qiu2025diaggym}. Dynamic case generation from
knowledge graphs avoids retrospective notes~\cite{clindef2025}. Closest to our
case-text finding, an audit of rare-disease diagnosis showed that LLMs infer diagnoses from retrospective therapeutic cues in clinical
narratives~\cite{rare2026audit}. We study common emergency diagnoses and quantify
how much each construction step contributes.
``Hindsight bias'' has been studied for the \emph{evaluated} model when it is shown future
data~\cite{hindsighttemporal2026}. We study hindsight in benchmark construction. Models are
known to exploit shortcuts in their data~\cite{geirhos2020shortcut}. In NLP, annotation artifacts and hypothesis-only baselines
showed that benchmarks can be solved from cues unrelated to the intended
reasoning~\cite{gururangan2018artifacts,poliak2018hypothesis,mccoy2019right}, and data
contamination is a recognized threat to the validity of LLM
evaluation~\cite{sainz2023contamination,golchin2024timetravel}. The case-text channel we study
resembles the former: the cue arises from how the evaluation input is built, not from
training data.

\section{The \method{} protocol}
\label{sec:protocol}

\begin{figure*}[t]
\centering
\resizebox{\ifdefined\ARXIV 0.95\fi\textwidth}{!}{\definecolor{ccbig}{HTML}{D55E00}   
\definecolor{ccfix}{HTML}{009E73}   
\begin{tikzpicture}[
  font=\small,
  box/.style={draw, rounded corners=2pt, align=center, minimum height=9mm, inner sep=3pt, fill=white},
  src/.style={box, fill=black!6},
  ch/.style={circle, draw, fill=black, text=white, inner sep=1.2pt, minimum size=4.6mm, font=\small\bfseries},
  lab/.style={font=\small, align=center, inner sep=1.5pt, fill=white},
  arr/.style={-{Latex[length=2.2mm]}},
]
  \node[src, text width=34mm] (ds) at (0,0.9) {Discharge summary\\(written after the diagnosis)};
  \node[src, text width=34mm] (ed) at (0,-1.85) {ED triage record\\(at arrival)};

  \node[box, text width=30mm, dashed] (dx) at (5.6,3.5) {Final diagnosis $D$};
  \node[box, text width=30mm, minimum height=13mm] (prof) at (5.6,1.9) {Simulator profile\\history $H$ $(+D)$};
  \node[box, text width=30mm] (case) at (5.6,0.0) {Agent-facing case\\(LLM summary or $H$)};
  \node[box, text width=30mm, draw=ccfix, very thick] (v3) at (5.6,-1.85) {Presentation-time\\reconstruction (V3)};

  \node[box, text width=20mm] (sim) at (10.6,1.9) {Simulator};
  \node[box, text width=22mm] (res) at (15.2,1.9) {Simulated lab results};
  \node[src, text width=22mm, minimum height=23mm] (agent) at (15.2,-0.93) {Diagnostic\\agent / LLM};

  \draw[arr, semithick, dashed, draw=black!60] (ds.east) -- ++(0.5,0) |- (prof.west);
  \draw[arr, line width=2.8pt, draw=ccbig] (ds.east) -- ++(0.5,0) |- (case.west);
  \draw[arr, thin, dashed, draw=black!55] (dx) -- (prof);
  \draw[arr, semithick] (prof) -- (sim);
  \draw[arr, semithick] (sim) -- (res);
  \draw[arr, semithick] (res) -- (agent);
  \draw[arr, line width=2.8pt, draw=ccbig] (case.east) -- (case.east -| agent.west);
  \draw[arr, semithick, draw=ccfix] (ed.east) -- (v3.west);
  \draw[arr, semithick, draw=ccfix, dashed] (ds.south) -- (ds.south |- 0,-1.15) -| ([xshift=-9mm]v3.north);
  \draw[arr, semithick, draw=ccfix] (v3.east) -- (v3.east -| agent.west);

  \node[ch] (c1) at ($(dx.north east)+(0.05,0.05)$) {1};
  \node[lab, anchor=west, align=left] at ($(c1.east)+(0.06,0)$) {diagnosis line:\\\textbf{negligible} ($\le$1.4 pp) in 5 simulators};
  \node[ch] (c2) at ($(prof.north east)+(0.05,0.05)$) {2};
  \node[lab, anchor=south west, align=left] (l2) at ($(c2.south east)+(0.06,-0.05)$) {retrospective history:\\small trace (+0.01--0.06)};
  \node[lab, anchor=south] at ($(sim.east)!0.5!(res.west)+(0,0.2)$) {gatekeepers keep\\16--62\% of\\real signal};
  \node[ch] (c3) at ($(case.north east)+(0.05,0.05)$) {3};
  \node[lab, text=ccbig, anchor=south west, align=left] at ($(c3.south east)+(0.06,-0.05)$)
        {LLM top-1: \textbf{+14 pp} (summary), \textbf{+20 pp} (history)\\six agents: \textbf{+7 pp} (summary), \textbf{+13 pp} (history)};
  \node[lab, text=ccbig, anchor=north] at ($(case.east)!0.5!(case.east -| agent.west)+(0,-0.25)$)
        {unchanged after masking diagnosis names};
  \node[lab, text=ccfix, anchor=south] at ($(v3.east)!0.5!(v3.east -| agent.west)+(0,0.2)$)
        {25\% top-1, same as chief complaint alone};
  \node[lab, text=ccfix, anchor=south] at (2.2,-1.15) {pre-admission sections only};
\end{tikzpicture}}
\caption{Three hindsight channels in a benchmark built from discharge summaries, with the measured
effect of each (dashed thin lines: little or no effect, thick line: large effect). The case text
\protect\chn{3} makes LLM diagnosis much easier than a presentation-time reconstruction (green), an
upper bound on hindsight (\S\ref{sec:v3}).}
\label{fig:channels}
\end{figure*}
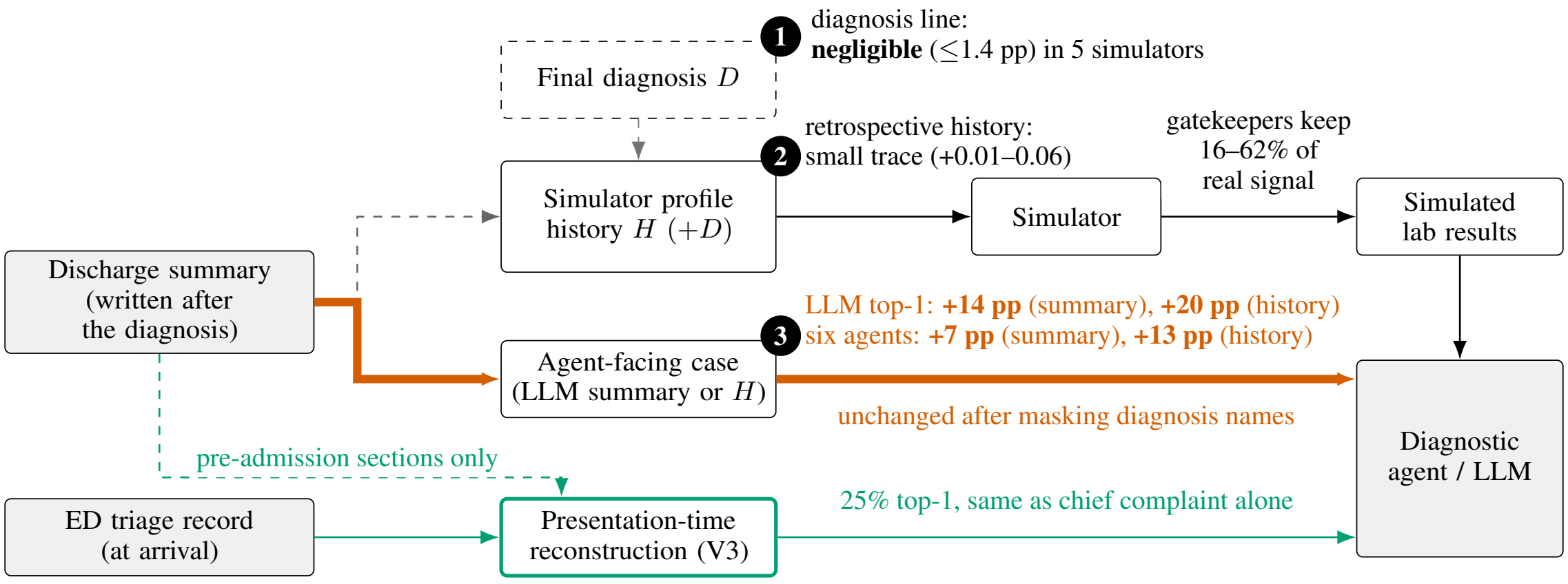

\paragraph{Channels.}
Let $D$ be the final diagnosis, $H$ the retrospective history (the discharge-summary text
preceding the physical examination section, as used by DiagGym), and $P$ the minimal
presentation-time information (sex and chief complaint). A simulator generates a result
$\hat{x}$ for an examination given a profile. We compare three profiles:
\splus{} $=(H,D)$, the standard benchmark setting, \sminus{} $=H$, and \szero{} $=P$.
The contrast \splus{}$-$\sminus{} isolates the diagnosis line \chn{1}, \sminus{}$-$\szero{} the
retrospective context \chn{2}, and \splus{}$-$\szero{} is the total simulator-side hindsight. The
agent-facing case text \chn{3} is compared with a richer presentation-time reconstruction than
$P$ (Figure~\ref{fig:channels}).

\paragraph{Case-text channel.} We compare benchmark-style texts (an LLM-written pre-hospital
summary and the discharge history) with presentation-time texts (V3, built from the triage
record and pre-admission history, and $P$) for the same admission. Zero-shot LLMs and text
classifiers trained on real data read each text, and the accuracy gap between texts measures
what the construction adds (\S\ref{sec:setup}).

\paragraph{Judges trained on real data only.}
For the simulator channels, leakage is measured as how well a fixed judge, trained only on real
training data, recovers $D$ from real or simulated laboratory
results. The primary judge
(\emph{num}) is a multinomial logistic regression over values, abnormal flags and missingness of
the 300 most common laboratory items, with lenient numeric parsing. Cross-simulator comparisons
use a \emph{values-only} variant, as DiagGym writes no abnormal flags, and a
gradient-boosted judge checks that conclusions do not hinge on a linear model. A text-and-value
judge (\emph{clf}) was primary in the first two registered tests, and a local LLM judge serves as
a reference. Diagnosis keywords are masked in
all results, real or simulated.

\paragraph{Metrics.}
For each case we compute the probability the judge assigns to the true category
(\ptrue{}), averaged over five simulator samples, plus top-1/top-5 accuracy. Contrasts are
paired by patient, and accuracy differences are given in percentage points (pp). Uncertainty is
given as 95\% percentile intervals from a patient-level bootstrap. \emph{Signal retention} is
$(\bar p_{\text{sim}}-c)/(\bar p_{\text{real}}-c)$, where $\bar p$ is the mean \ptrue{} of simulated or
real results and $c=1/50$ is chance (1: as informative as real results). \emph{Typicalization} is
the fraction of real results that are atypical for the diagnosis (category-item abnormal
rate $\ge 0.6$ but real result normal, or $\le 0.2$ but real abnormal) that a simulator
``corrects'' to the typical state.

\paragraph{Sequential registration.}
Before each run, the hypothesis, primary metric, decision rule and action on failure were
written into a timestamped, versioned protocol and not changed afterwards. Registration was
\emph{sequential} and internal: each test was designed after the previous result, so the record
documents what was decided and when, not a single pre-specified plan. We label the registered tests D1, A, A2, A$'$1--A$'$3 and E1 in the
order they were run (Table~\ref{tab:prereg}). The tests A$'$1--A$'$3 form one family with Holm
correction~\cite{holm1979}, and the replication E1 was registered after the exploratory
case-text analyses. Error control is claimed only within these sets, and the one positive registered result
(E1) has its own Holm correction. Rules, timing and the protocol history are
deposited at OSF (Appendix~\ref{app:history}).

\section{Data and experimental setup}
\label{sec:setup}

\paragraph{Cohort.}
We use MIMIC-IV v3.1 with MIMIC-IV-Note v2.2~\cite{johnson2023mimiciv,johnson2024mimiciv31,johnson2023mimicnote}. Principal
diagnoses are mapped to AHRQ CCSR categories~\cite{ahrq2026ccsr} (ICD-9 via CMS GEMs). We keep emergency or
urgent admissions whose discharge summary contains a history, a physical examination and a
discharge diagnosis section (264{,}058 cases). We exclude injury (INJ) and mental and behavioral
(MBD) categories, as well as transfers and post-procedure admissions, whose diagnosis is often
established before arrival. We keep one admission per patient, take the 50 most frequent
categories with at most 600 cases each, and split the data 8:1:1 by patient. Following
DiagGym, examination events are taken from 24\,h before to 24\,h after admission, keeping
the first occurrence of each. Admission physical examinations are structured by a local
Qwen2.5-7B-Instruct~\cite{qwen2025} model (98.6\% parse success), and cases with only a
discharge examination or a diagnosis judged known before admission are removed, leaving
\textbf{27{,}489 cases} (21{,}994 / 2{,}748 / 2{,}747 for training, validation and test). Simulator
analyses use 460 test cases (about 10 per category).
All models run locally (Appendix~\ref{app:compute}), and differences from DiagGym's pipeline are
listed in Appendix~\ref{app:diff}.

\paragraph{Simulators.}
\textbf{FT-3B}: Qwen2.5-3B-Instruct with LoRA~\cite{hu2021lora} trained on the DiagGym
text format with diagnosis dropout 0.5, so the same model runs as \splus{} or \sminus{}.
\textbf{GK-14B}: Qwen2.5-14B-Instruct, AWQ-quantized~\cite{lin2024awq}, prompted as an SDBench-style gatekeeper.
\textbf{GK-32B}: Qwen2.5-32B-Instruct-AWQ gatekeeper, additionally told the most common
real unit for each test. \textbf{DiagGym 7B}: the public weights, used with its own
inference format. \textbf{Phi-4}~\cite{abdin2024phi4}: a 14B model from a different family, prompted exactly
like GK-32B. \textbf{GK-72B}: Qwen2.5-72B-Instruct-AWQ, prompted like GK-32B. An unquantized (BF16) copy of
GK-14B serves as a quantization check. Results are sampled five
times per item at temperature 1.0, either conditioned on preceding real results
(\emph{sequential}) or from the profile alone (\emph{independent}). The registered tests used FT-3B, GK-14B and GK-32B. The others are
descriptive replications. DiagGym was also run on 225 DiagBench cases its authors held out from
training (Appendix~\ref{app:simextra}).

\paragraph{Agent-facing views and LLM diagnosis.}
For the case-text analyses (\S\ref{sec:v3}) we compare four views of the same admission:
\textbf{V1}, a DiagGym-style pre-hospital summary written by Qwen2.5-14B-Instruct-AWQ with
DiagGym's verbatim instruction and kept only if it passes DiagGym's leakage filter. \textbf{V2}, the discharge-summary history (DiagGym's simulator
context). \textbf{V3}, a presentation-time reconstruction: sex, the MIMIC-IV-ED~\cite{johnson2023mimiced} triage record
of the visit that led to the admission (chief complaint, vital signs, pain score, acuity) and
the discharge-summary sections describing the patient \emph{before} the admission
(allergies, past medical history, family history), excluding the history of present illness
and the procedure section. Because these sections come from the discharge summary, 32\% of V3
texts name a diagnosis term (53\% of histories, 19\% of chief complaints), partly as legitimately
known chronic conditions. \textbf{V4} is sex and chief complaint, i.e., $P$. Four local LLMs
(Qwen2.5-7B, -14B-AWQ and -32B-AWQ, and Phi-4, with Llama-3.1-8B-Instruct~\cite{grattafiori2024llama3} added afterwards as a third
family) read one view at temperature 0 and rank the
50 CCSR groups, given by code and description (prompts in Appendix~\ref{app:prompts}). Variants mask
diagnosis terms in every view or use a differently worded prompt, and the registered replication
E1 repeats the analysis on the validation split. A format control (V3n)
rewrites V3 as a narrative with the same model and instruction that produced V1. A TF-IDF + logistic
regression classifier per view (scikit-learn~\cite{pedregosa2011sklearn}) provides a non-LLM reference. Zero-shot top-1 over 50 groups
is a probe of how much diagnostic information a text carries, not a measure of clinical
ability.

\paragraph{Agent.}
For the end-to-end test, a Qwen2.5-14B-Instruct-AWQ agent (temperature 0) sees sex and
chief complaint, may order up to five panels (blood count, chemistry, liver panel,
lactate, urinalysis), and returns a top-5 list over the 50 categories. Panel results come
from real replay or from GK-14B (independent generation with unit hints) given \splus{} or
\szero{}. Cases are the 721
test cases with all five panels recorded. As descriptive checks, a stronger
Qwen2.5-32B-Instruct-AWQ agent was paired with the 32B gatekeeper, and both agents were also run
with no laboratory results.

\section{Results}

\subsection{Retrospective case text carries the answer}
\label{sec:casetext}

\begin{table}[t]
\centering

\small
\setlength{\tabcolsep}{3pt}
\begin{tabular}{@{}lccc@{}}
\toprule
Input & top-1 & top-5 & \ptrue{} \\
\midrule
Retrospective history $H$             & 0.559 & 0.891 & -- \\
$H$ with diagnosis words deleted      & 0.558 & 0.878 & -- \\
Chief complaint                       & 0.368 & 0.702 & -- \\
Sex + chief complaint ($P$)           & 0.366 & 0.703 & -- \\
\midrule
LLM pre-hospital summary (filtered)   & 0.491 & 0.857 & 0.192 \\
Chief complaint (same cases)          & 0.357 & 0.706 & 0.190 \\
\bottomrule
\end{tabular}
\caption{Diagnostic predictability of agent-facing text (TF-IDF + logistic regression
trained on the training split). Top: validation split, 2{,}748 cases. Bottom: test cases
whose DiagGym-style summary passed the explicit-leakage filter, 1{,}760 cases.}
\label{tab:upstream}
\end{table}

The retrospective history alone predicted the diagnosis in 55.9\% of cases, more than a judge
reading all real examination results (40.0\%), and deleting diagnosis words did not reduce it
(Table~\ref{tab:upstream}). 52\% of histories mention a diagnosis word and 42--47\% list a
performed procedure. The profile given to \sminus{} and \splus{} simulators therefore already
contains much of the answer. The gain over the chief complaint was broad (43 of 50 categories,
Appendix~\ref{app:casetext}) but partly legitimate, because known conditions such as cancer are
part of the past medical history at presentation.

DiagGym-style pre-hospital summaries, generated with its instruction and filtered with its
leakage criterion by a local 14B model, were rejected for stating the diagnosis in 34.6\% of
cases. Among those that passed, 46.9\% still contained a diagnosis word, and top-1
predictability rose from 35.7\% to 49.1\% ($+13.4$ pp [$+10.9$, $+15.9$]). The registered primary
metric of this test (A$'$1), \ptrue{}, showed no difference ($+0.002$, $p=0.32$) because it
compared two differently calibrated classifiers. Following our rules, we report A$'$1 as not
supported and the accuracy contrast as secondary (Appendix~\ref{app:calib}).

\subsection{Presentation-time reconstruction and LLM diagnosis (exploratory)}
\label{sec:v3}

The presentation-time reconstruction V3 narrows the upper bound left by the chief complaint but
does not close it: it lacks the history of present illness taken at arrival, because MIMIC-IV
contains no contemporaneous emergency-physician note. The V1$-$V3 and V2$-$V3 contrasts therefore
measure how much easier benchmark text makes diagnosis than the structured presentation-time
record, an upper bound on hindsight. We analyze the 1{,}053 test cases with an ED triage record
(about 60\% of admissions) and a summary that passed the leakage filter. Restricting to
triage-linked admissions did not change the history-versus-complaint gap (validation split:
+19.3 vs.\ +20.2 pp). These analyses are exploratory.

\begin{table}[t]
\centering
\small
\setlength{\tabcolsep}{2.4pt}
\begin{tabular}{@{}lrrrrrr@{}}
\toprule
 & V1 & V2 & V3 & V4 & V1$-$V3 & V2$-$V3 \\
\midrule
Qwen2.5-7B   & 37.5 & 42.8 & 23.1 & 23.2 & +14.4 & +19.8 \\
Qwen2.5-14B  & 39.3 & 45.1 & 24.4 & 24.5 & +14.9 & +20.7 \\
Qwen2.5-32B  & 40.0 & 46.0 & 27.5 & 25.8 & +12.4 & +18.4 \\
Phi-4        & 41.9 & 47.2 & 26.7 & 27.3 & +15.2 & +20.5 \\
Llama-3.1-8B$^\dagger$\! & 34.4 & 39.8 & 17.7 & 18.9 & +16.7 & +22.1 \\
\midrule
Mean (four)  & 39.7 & 45.3 & 25.4 & 25.2 & +14.2 & +19.8 \\
\quad masked & 36.1 & 41.7 & 21.4 & 20.1 & +14.8 & +20.3 \\
14B, alt.\ prompt & 29.1 & 26.2 & 19.1 & 22.1 & +10.0 & +7.1 \\
\midrule
E1, validation & 36.2 & 43.1 & 25.0 & 23.5 & +11.1 & +18.1 \\
\midrule
Text classifier & 48.1 & 55.1 & 33.1 & 35.0 & +14.9 & +21.9 \\
\bottomrule
\end{tabular}
\caption{Top-1 accuracy (\%) of zero-shot diagnosis (50 CCSR groups) from four views of the same
1{,}053 test cases: V1 benchmark summary, V2 discharge history, V3 presentation-time
reconstruction, V4 sex + chief complaint. Differences in percentage points, computed from unrounded accuracies. Patient-level
bootstrap 95\% CIs are about $\pm$3 pp per model and $\pm$2.3 pp for the mean (all $p<0.001$).
\emph{Masked}: diagnosis terms deleted from every view (mean of the four models).
$^\dagger$Added afterwards and not included in the means. \emph{E1}: registered replication on the 1{,}091
validation cases (\S\ref{sec:v3}).}
\label{tab:llmdx}
\end{table}

\paragraph{Benchmark-style inputs make LLM diagnosis much easier.} Reading the summary, four
LLMs diagnosed correctly 14.2 pp more often than from V3 (95\% CI [12.0, 16.5], per model
12.4--15.2 pp), and reading the discharge history 19.8 pp more often [17.5, 22.2]
(Table~\ref{tab:llmdx}), a relative increase of more than 50\%. A text classifier showed the same
pattern (+14.9 and +21.9 pp), and Llama-3.1-8B, from a third family, showed the largest gaps
(+16.7 and +22.1 pp).

\begin{figure}[t]
\centering
\includegraphics[width=\columnwidth]{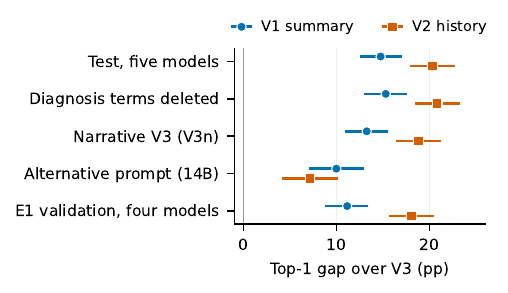}
\caption{Top-1 gap of zero-shot LLM diagnosis between the benchmark text and the
presentation-time view (V3n: V3 rewritten as a narrative) under each condition, mean over the
models listed, with 95\% patient-level bootstrap CIs. E1 is the registered replication. Per-model
estimates are in Figure~\ref{fig:forest} (Appendix~\ref{app:casetext}).}
\label{fig:forestmean}
\end{figure}

\paragraph{The gap replicates on held-out patients.} We registered a replication (E1) before
running it: the same four models, prompt and views on the validation split, whose patients do
not overlap the test split and on which no LLM analysis had been run. For both four-model gaps,
the null hypothesis had to be rejected under Holm correction and the gap had to be at least 5 pp. On 1{,}091 validation cases,
V1$-$V3 was +11.1 pp [8.9, 13.2] and V2$-$V3 +18.1 pp [15.8, 20.3] (both $p<0.001$,
Figure~\ref{fig:forestmean}), so the rule was met. Both splits come from the same pilot cohort,
so this is a replication on held-out patients, not an independent confirmatory study.

\paragraph{The gap is not a format effect.} Rewriting V3 as a narrative with the model and
instruction that produced V1 (V3n) raised accuracy by only 1.5 pp [0.5, 2.5], leaving V1$-$V3n at
+13.2 pp [11.1, 15.4] and V2$-$V3n at +18.8 pp [16.6, 21.1] (five models). The rewriting named a
diagnosis term more often than V3 itself (531 vs.\ 450 of 1{,}635 cases), so this control, if
anything, favors V3n.

\paragraph{The gap lies beyond diagnosis names.} Deleting diagnosis terms from every view lowered
all accuracies by 3--5 pp but left the gap unchanged (V1$-$V3 +14.8 pp [12.5, 17.1], V2$-$V3
+20.3 pp [18.0, 22.7], four-model mean). Deleting classes of words from the summaries and
refitting the text classifier changed top-1 by less than 1 pp for diagnosis terms or for
work-up, treatment, hospital-course or history words. Deleting all of them together still left
0.499, 14 pp above the chief complaint (0.358, Appendix~\ref{app:cues}). The most weighted
words are symptom descriptions, synonyms and abbreviations that a name list misses, such as
\emph{melena} for gastrointestinal hemorrhage or \emph{flatus} for intestinal obstruction. Some
of these could have been elicited at the bedside, and the text alone cannot show whether they
were reported at arrival or selected by an author who knew the answer. In either case, they form
a lexical shortcut that name filters cannot remove and that lets a model score by matching words
rather than by clinical reasoning.

\paragraph{Structured presentation-time information adds little.} Triage vital signs and
pre-admission history did not improve on the chief complaint (V3$-$V4 +0.2 pp [$-2.2$, 2.8]).
The gap persisted, smaller, under a differently worded prompt (+10.0 pp for V1) and appeared in
82\% of the 49 categories with at least ten cases (Appendix~\ref{app:casetext}).

\paragraph{Interactive agents show the same gap, but keep their ranking.} Six local agents (7B to
72B parameters, three families) read V1, V2 or V3 of the same 1{,}053 cases, ordered up to eight
examinations with real results (laboratory panels, the first imaging report, the admission
examination), and then diagnosed. Every agent scored higher from the benchmark text, on average by
7.2 pp [5.6, 8.8] for V1 and 12.5 pp [10.8, 14.2] for V2, but the ranking of the agents barely
changed (Kendall $\tau=0.87$ between V1 and V3; Appendix~\ref{app:agentviews}). Hindsight in the case
text thus inflates absolute agent scores more than it reorders agents, and such scores should not be read as bedside accuracy.

\subsection{Registered simulator and agent tests}
\label{sec:prereg}

Table~\ref{tab:prereg} lists every registered test in the order it was run. Each
failure led to a design revision whose rationale was recorded before the next run
(Appendix~\ref{app:history}).

\begin{table*}[t]
\centering

\small
\begin{tabular}{@{}llllc@{}}
\toprule
Test & Simulator / setting & Primary metric & $\Delta$ [95\% CI] & Met? \\
\midrule
D1 & FT-3B, sequential & clf top-1, \splus{}$-$real & $-0.080$ [$-0.113$, $-0.046$] & no \\
A  & GK-14B, sequential & clf top-1, \splus{}$-$\sminus{} & $+0.007$ [$-0.004$, $+0.017$] & no \\
A2 & GK-14B, independent & \makecell[l]{num \ptrue{}, \splus{}$-$real\\ \quad and \splus{}$-$\szero{}} &
  \makecell[l]{$-0.143$ [$-0.172$, $-0.114$]\\ $+0.017$ [$+0.007$, $+0.027$]} & no \\
\midrule
\multicolumn{5}{@{}l}{\emph{Family A$'$ (one-sided, Holm, $\alpha=0.05$)}}\\
A$'$1 & summary vs.\ chief complaint & text clf \ptrue{}, summary$-$complaint & $+0.002$ [$-0.008$, $+0.013$], $p=0.320$ & no \\
A$'$2 & agent, GK-14B & top-1, \splus{}$-$\szero{} & $+0.000$ [$-0.015$, $+0.015$], $p=0.538$ & no \\
A$'$3 & GK-32B + units, independent & num \ptrue{}, \splus{}$-$real & $-0.117$ [$-0.146$, $-0.089$], $p\approx1$ & no \\
\midrule
\multicolumn{5}{@{}l}{\emph{Replication of the exploratory case-text result (one-sided, Holm over two, $\alpha=0.05$)}}\\
E1 & LLM diagnosis, validation & top-1, V1$-$V3 and V2$-$V3 & \makecell[l]{$+0.111$ [$+0.089$, $+0.132$], $p<0.001$\\ $+0.181$ [$+0.158$, $+0.203$], $p<0.001$} & yes \\
\bottomrule
\end{tabular}
\caption{All registered tests. $\Delta$ is the primary contrast with 95\% bootstrap CI and,
for family A$'$, the one-sided $p$. Top-1 contrasts are differences in proportion (0.010 = 1
percentage point). Five of the six tests above the last rule concern the simulator and A$'$1 the agent-facing
summary. None met its decision rule. E1, registered after the exploratory case-text analyses,
met its rule (null rejected for both gaps, both $\ge$0.050).}
\label{tab:prereg}
\end{table*}

\paragraph{Knowing the diagnosis does not make simulated results more diagnostic.} In every
setting, \splus{} results were \emph{less} predictable than real results
(Table~\ref{tab:prereg}). The diagnosis line added at most 1.35 pp (FT-3B, \splus{}$-$\sminus{}
top-1 [$+0.30$, $+2.39$]) and nothing measurable for the gatekeepers (GK-14B: \ptrue{} $-0.0005$
[$-0.007$, $+0.007$]). The descriptive replications agree for DiagGym, Phi-4 and GK-72B
(Table~\ref{tab:simulators}). This null does not reflect an insensitive judge, because the same
judges detect the retrospective-context trace and the gatekeeper dilution (\S\ref{sec:retention}).

\paragraph{The agent test was inconclusive overall.} The 14B agent reached top-1 of 20.7\% with
real results and 19.3\% with either gatekeeper (\splus{}$-$real $-1.39$ pp [$-3.33$, $+0.42$]),
but also 19.6\% with no laboratory results at all, so it made little use of them. A stronger 32B
agent gained only 1.9 pp [0.0, 3.9] from real results. The registered agent test (A$'$2) therefore
cannot show whether
simulator leakage matters for an agent that uses the results.

\paragraph{Where agents use laboratory results, the simulator matters (exploratory).} Many of the
50 categories are usually established by imaging or procedures, about which the five panels are
uninformative, so the full agent test is dominated by cases in which laboratory results matter
little. In a post hoc subset of eight categories usually established by the five panels (175 cases, with the list
fixed before per-category agent results were inspected, see Appendix~\ref{app:simextra}), the 32B
agent did use the results (real $-$ none +4.0 pp [0.6, 8.0]), and its accuracy depended on the
simulator: \splus{}$-$\szero{} was +4.0 pp [1.1, 7.4] in top-1 and +9.1 pp [1.7, 16.6] in top-5,
and \splus{}$-$real +8.0 pp [2.3, 13.7] in top-5, although for a real-data judge the same
simulated results carry less signal than real ones (\ptrue{} 0.110 vs.\ 0.264). The 14B agent,
which did not use the results, showed no such differences. These contrasts are exploratory,
uncorrected for multiplicity, and confound the diagnosis line with the retrospective profile.
Interpretive wording does not explain them (0.4\% of results under both \splus{} and \szero{}).
A plausible source is typicalization, since the 32B gatekeeper turned 61\% of atypically normal
real results into the typical abnormal state when given the retrospective profile and the
diagnosis (\splus{}), and 39\% when given only the chief complaint (\szero{}).

\subsection{Prompted gatekeepers dilute diagnostic signal, unlike trained world models}
\label{sec:retention}

\begin{table*}[t]
\centering
\small
\setlength{\tabcolsep}{2.8pt}
\begin{tabular}{@{}lllccccc@{}}
\toprule
Simulator & Type & Gen. & \splus{}$-$\sminus{} & \splus{}$-$\szero{} & Retention, linear & Retention, boosted & Typ.\ \splus{}/\szero{} \\
\midrule
FT-3B (ours)$^*$ & trained & seq. & +0.001 & -- & 0.84 [0.75, 0.94] & 0.81 [0.73, 0.89] & -- \\
DiagGym 7B & trained & seq. & +0.007 & -- & 0.77 [0.67, 0.88] & 0.82 [0.74, 0.92] & -- \\
GK-14B$^*$ & prompted & seq. & $-$0.001 & -- & 0.25 [0.14, 0.37] & 0.46 [0.37, 0.55] & -- \\
GK-14B$^*$ & prompted & ind. & +0.002 & +0.019 [+0.011, +0.028] & 0.16 [0.06, 0.27] & 0.43 [0.34, 0.53] & 0.45/0.36 \\
GK-32B + units$^*$ & prompted & ind. & -- & +0.028 [+0.015, +0.042] & 0.32 [0.21, 0.46] & 0.58 [0.50, 0.67] & 0.61/0.39 \\
GK-72B + units & prompted & ind. & +0.003 & +0.023 [+0.012, +0.035] & 0.27 [0.17, 0.40] & 0.62 [0.53, 0.72] & 0.61/0.53 \\
Phi-4 + units & prompted & ind. & +0.000 & +0.014 [+0.003, +0.025] & 0.20 [0.10, 0.31] & 0.49 [0.41, 0.57] & 0.62/0.48 \\
DiagGym 7B & trained & ind. & +0.004 & +0.059 [+0.042, +0.077] & 0.72 [0.60, 0.86] & 0.76 [0.67, 0.85] & -- \\
DiagGym, held-out & trained & ind. & +0.007 & +0.052 [+0.032, +0.074] & 0.70 [0.55, 0.88] & 0.75 [0.65, 0.86] & -- \\
\bottomrule
\end{tabular}

\caption{Overview of the simulators. Seq.: conditioned on the real earlier results.
Ind.: generated from the profile alone (the two are not comparable). Contrasts are in
\ptrue{} under the values-only linear judge. \splus{}$-$\sminus{} is a point estimate (paired
intervals for the registered tests are given in the text). Retention is computed as in
Figure~\ref{fig:retention}, with 95\% patient-level bootstrap CIs. $^*$Used in a registered
test. All other rows are descriptive. --: arm not generated.}
\label{tab:simulators}
\vspace{-3pt}
\end{table*}
Two kinds of simulator behaved very differently (Table~\ref{tab:simulators},
Figure~\ref{fig:retention} in Appendix~\ref{app:simextra}). With the values-only judge, independent \splus{}
results from \emph{prompted gatekeepers} retained 16--32\% of the real signal above chance
(43--62\% with the boosted judge), against 72\% for the \emph{trained world model} DiagGym. In
sequential generation, the trained models retained 77--84\% and GK-14B 25\%. Our FT-3B and the public DiagGym are thus close, and
the latter showed no sign of memorizing our cases (Appendix~\ref{app:simextra}). The magnitude depends
on the judge, but the ordering does not. Retention rose from GK-14B to GK-32B, which also adds
unit hints, but not further from GK-32B to GK-72B, which share them, so dilution did not
disappear with scale. Its degree may still reflect the
limits of the open models we could run locally and should not be
extrapolated to frontier models. Quantization does not explain it: an unquantized (BF16) copy of
GK-14B retained 19\% [9, 30] of the signal (boosted judge 47\%), against 16\% and 43\% for the
quantized version (Appendix~\ref{app:simextra}). The gap was not a
formatting artifact (Appendix~\ref{app:units}).
Retention measures the information in simulated results, not changes in agent scores.

\paragraph{Hindsight leaves a small, consistent trace.} \splus{}$-$\szero{} was positive for every
simulator (+0.014 to +0.059 in \ptrue{}, largest for DiagGym, Table~\ref{tab:simulators}), and
gatekeepers typicalized more with the retrospective profile. Because \splus{} and \sminus{} were
indistinguishable, the trace comes from the retrospective history, not from the diagnosis line.
These contrasts are secondary.

\section{Discussion}
\label{sec:discussion}

\paragraph{Where the leakage lies.} Benchmark builders invest in
hiding the diagnosis from simulators and filtering diagnosis names from agent inputs. Our
results suggest that the first measure addresses a small effect, while the second leaves
untouched most of the extra diagnostic signal that retrospective text carries.

\paragraph{How firm is the case-text evidence?} The central result was found exploratorily,
after the registered simulator tests had failed, but it passed a registered replication on
held-out patients and holds across five models from three families, a non-LLM classifier, a
narrative rewriting of the presentation-time view, masking by placeholder or deletion, an
alternative prompt, 82\% of categories and a sensitivity analysis (Appendix~\ref{app:sens}). It
cannot establish how much of the gap is hindsight rather than history a clinician could have
taken at the bedside. That requires contemporaneous admission notes, which MIMIC lacks.

\paragraph{The choice of simulator matters more than what it is told.} Whether the simulator
sees the diagnosis made little difference (at most 1.4 pp), but whether it was a prompted
gatekeeper or a trained world model changed signal retention substantially (16--32\% vs.\ 72\%
with the values-only judge). The exploratory agent subset suggests that a gatekeeper holding the
full case file, including the diagnosis, can inflate the accuracy of an agent that uses its
results, even though these results carry less signal than real ones. Whether simulators change
agent scores and rankings is the key open question.

\paragraph{Two pitfalls for leakage audits.} Masking diagnosis terms with a placeholder gave
text classifiers a new cue (summary top-1 0.525 vs.\ 0.491 unmasked), whereas deleting the terms
did not. Comparing raw probabilities across differently calibrated classifiers hid a
13.4-pp accuracy gain in A$'$1. Audits should mask by deletion and compare classifiers by
accuracy or proper scoring rules (Appendix~\ref{app:pitfalls}).

\paragraph{Recommendations.} Benchmarks should build agent-facing cases from information
documented at presentation, or report a presentation-time baseline next to every score. They
should not rely on diagnosis-name filters, should report the signal retention of simulated
results against real results, and should verify that agents use the simulated results before
attributing score differences to simulators. Table~\ref{tab:checklist} (Appendix~\ref{app:recs}) links each
recommendation to its evidence.

\section{Conclusion}
Telling simulators the diagnosis made simulated results at most marginally more diagnostic
(at most 1.4 pp) for judges trained on real data, and never more diagnostic than real results.
Prompted gatekeepers retained only a fraction of the real signal, whereas trained world models
retained most of it, and simulators may still affect the scores of agents that rely on
examination results. The case text made diagnosis markedly easier than the structured
information available at presentation, for zero-shot models and interactive agents alike, in ways
that diagnosis-name filters cannot remove.
Benchmarks should therefore be built from, or at least compared against, presentation-time
information. How much of this gap reflects hindsight, rather than
history a clinician could obtain at the bedside, can only be resolved with contemporaneously
written admission notes.

\section*{Limitations}
\paragraph{Data and models.} We use a single data source (MIMIC-IV, one US center), and all
models were served by local inference. The simulators span two model families (Qwen2.5,
including DiagGym, which is fine-tuned from it, and Phi-4), and the study includes no proprietary
model such as the o4-mini gatekeeper of SDBench. DiagGym's training split is unpublished, so our
memorization check relies on its authors' held-out DiagBench cases (225 usable) and compares
different case sets. The DiagGym, Phi-4, GK-72B, BF16 and Llama-3.1-8B analyses were added after
the registered tests and are descriptive.

\paragraph{Measurement.} The size of the dilution depends on the judge, because a boosted judge
recovered about two to three times the signal of a linear one, so we report ranges rather than a single number. Judges stronger than ours might detect simulator leakage that ours missed. The simulator audit covers laboratory panels only. Imaging, pathology and endoscopy
reports are free text that a gatekeeper aware of the diagnosis could phrase in ways that point to
it, so the simulator channels may be larger for such examinations.

\paragraph{Case-text analyses.} The presentation-time analyses were found exploratorily on the
pilot test split and replicated on its validation split, which belongs to the same cohort. They
cover the 60\% of admissions with an ED triage record. The triage chief complaint is terser than
the discharge-summary one, social history is de-identified, and medication reconciliation was
not included, so V3 may understate legitimately available information. V3n matches the narrative format of V1 but not its length or the amount of information it carries. Above all, V3 has no
history of present illness, because MIMIC-IV has no contemporaneous emergency-physician note.
The V1$-$V3 and V2$-$V3 contrasts, like the history-versus-complaint contrasts, are therefore
upper bounds on hindsight rather than estimates of it. The size of the LLM gap was
prompt-sensitive, and only one alternative prompt was tested.

\paragraph{Agents and registration.} The A$'$2 agents were weak (top-1 about 20--24\%, gaining at most
2 pp from laboratory results across all cases) and could order only five panels, which limits the power of A$'$2.
As noted in \S\ref{sec:casetext}, the primary metric of A$'$1 was poorly chosen. The registration was
internal, timestamped and sequential, with each test designed after the previous result, rather
than a single plan deposited in a public registry beforehand.

\paragraph{Scope.} We studied 50 common emergency categories, excluding injury and mental
disorders, and all three channels may behave differently for rare diseases. Six interactive agents
kept nearly the same order on V3 as on the benchmark text, but whether presentation-time benchmarks
separate agents of similar strength remains untested.

\section*{Ethics Statement}
All data come from MIMIC-IV, MIMIC-IV-Note and MIMIC-IV-ED, used under the PhysioNet~\cite{goldberger2000physionet}
credentialed data use agreement. Every language model ran locally on a server accessible only to credentialed project members.
No record-level text was sent to an online service, and only aggregate statistics left the
server. We do not release patient-level text or generated records. Code and aggregate results
will be released. The work audits evaluation practice and does not deploy any clinical system.

\bibliography{refs}

\begin{thebibliography}{29}
\providecommand{\natexlab}[1]{#1}

\bibitem[{Abdin et~al.(2024)Abdin, Aneja, Behl, Bubeck, Eldan, Gunasekar
  et~al.}]{abdin2024phi4}
Marah Abdin, Jyoti Aneja, Harkirat Behl, S{\'e}bastien Bubeck, Ronen Eldan,
  Suriya Gunasekar, et~al. 2024.
\newblock {Phi-4} technical report.
\newblock \emph{arXiv preprint arXiv:2412.08905}.

\bibitem[{{Agency for Healthcare Research and Quality}(2026)}]{ahrq2026ccsr}
{Agency for Healthcare Research and Quality}. 2026.
\newblock Clinical classifications software refined ({CCSR}) for {ICD-10-CM}
  diagnoses, v2026.1.
\newblock Healthcare Cost and Utilization Project (HCUP).
  \url{https://hcup-us.ahrq.gov/toolssoftware/ccsr/dxccsr.jsp}.

\bibitem[{Dutta et~al.(2026)Dutta, Mishra, Singh, and Moharir}]{pulsecare2026}
Srimonti Dutta, Dinky Mishra, Aditi Singh, and Akshata~Kishore Moharir. 2026.
\newblock \href {https://aclanthology.org/2026.aacl-long.266/} {{PULSE}-{CARE}:
  A measurement-validity audit of simulated-patient clinical agent evaluation}.
\newblock In \emph{Proceedings of IJCNLP-AACL 2026 (Volume 1: Long Papers)},
  pages 5904--5923. Association for Computational Linguistics.

\bibitem[{Geirhos et~al.(2020)}]{geirhos2020shortcut}
Robert Geirhos et~al. 2020.
\newblock Shortcut learning in deep neural networks.
\newblock \emph{Nature Machine Intelligence}, 2:665--673.

\bibitem[{Golchin and Surdeanu(2024)}]{golchin2024timetravel}
Shahriar Golchin and Mihai Surdeanu. 2024.
\newblock Time travel in {LLM}s: Tracing data contamination in large language
  models.
\newblock In \emph{Proceedings of ICLR}.
\newblock ArXiv:2308.08493.

\bibitem[{Goldberger et~al.(2000)Goldberger, Amaral, Glass, Hausdorff, Ivanov,
  Mark, Mietus, Moody, Peng, and Stanley}]{goldberger2000physionet}
Ary~L. Goldberger, Luis A.~N. Amaral, Leon Glass, Jeffrey~M. Hausdorff,
  Plamen~Ch. Ivanov, Roger~G. Mark, Joseph~E. Mietus, George~B. Moody,
  Chung-Kang Peng, and H.~Eugene Stanley. 2000.
\newblock \href {https://doi.org/10.1161/01.CIR.101.23.e215} {{PhysioBank},
  {PhysioToolkit}, and {PhysioNet}: Components of a new research resource for
  complex physiologic signals}.
\newblock \emph{Circulation}, 101(23):e215--e220.

\bibitem[{Grattafiori et~al.(2024)Grattafiori, Dubey, Jauhri, Pandey, Kadian,
  Al-Dahle et~al.}]{grattafiori2024llama3}
Aaron Grattafiori, Abhimanyu Dubey, Abhinav Jauhri, Abhinav Pandey, Abhishek
  Kadian, Ahmad Al-Dahle, et~al. 2024.
\newblock The {Llama} 3 herd of models.
\newblock \emph{arXiv preprint arXiv:2407.21783}.

\bibitem[{Gururangan et~al.(2018)Gururangan, Swayamdipta, Levy, Schwartz,
  Bowman, and Smith}]{gururangan2018artifacts}
Suchin Gururangan, Swabha Swayamdipta, Omer Levy, Roy Schwartz, Samuel~R.
  Bowman, and Noah~A. Smith. 2018.
\newblock Annotation artifacts in natural language inference data.
\newblock In \emph{Proceedings of NAACL-HLT}.
\newblock ArXiv:1803.02324.

\bibitem[{Hager et~al.(2024)Hager, Jungmann, Holland, Bhagat, Hubrecht, Knauer,
  Vielhauer, Makowski, Braren, Kaissis, and Rueckert}]{hager2024mimiccdm}
Paul Hager, Friederike Jungmann, Robbie Holland, Kunal Bhagat, Inga Hubrecht,
  Manuel Knauer, Jakob Vielhauer, Marcus Makowski, Rickmer Braren, Georgios
  Kaissis, and Daniel Rueckert. 2024.
\newblock \href {https://doi.org/10.1038/s41591-024-03097-1} {Evaluation and
  mitigation of the limitations of large language models in clinical
  decision-making}.
\newblock \emph{Nature Medicine}, 30(9):2613--2622.

\bibitem[{Holm(1979)}]{holm1979}
Sture Holm. 1979.
\newblock A simple sequentially rejective multiple test procedure.
\newblock \emph{Scandinavian Journal of Statistics}, 6(2):65--70.

\bibitem[{Hu et~al.(2022)}]{hu2021lora}
Edward~J. Hu et~al. 2022.
\newblock {LoRA}: Low-rank adaptation of large language models.
\newblock In \emph{ICLR}.

\bibitem[{Johnson et~al.(2023{\natexlab{a}})Johnson, Bulgarelli, Pollard, Celi,
  Mark, and Horng}]{johnson2023mimiced}
Alistair Johnson, Lucas Bulgarelli, Tom Pollard, Leo~Anthony Celi, Roger Mark,
  and Steven Horng. 2023{\natexlab{a}}.
\newblock \href {https://doi.org/10.13026/5ntk-km72} {{MIMIC-IV-ED} (version
  2.2)}.
\newblock PhysioNet.

\bibitem[{Johnson et~al.(2024)Johnson, Bulgarelli, Pollard, Gow, Moody, Horng,
  Celi, and Mark}]{johnson2024mimiciv31}
Alistair Johnson, Lucas Bulgarelli, Tom Pollard, Brian Gow, Benjamin Moody,
  Steven Horng, Leo~Anthony Celi, and Roger Mark. 2024.
\newblock \href {https://doi.org/10.13026/kpb9-mt58} {{MIMIC-IV} (version
  3.1)}.
\newblock PhysioNet.

\bibitem[{Johnson et~al.(2023{\natexlab{b}})Johnson, Pollard, Horng, Celi, and
  Mark}]{johnson2023mimicnote}
Alistair Johnson, Tom Pollard, Steven Horng, Leo~Anthony Celi, and Roger Mark.
  2023{\natexlab{b}}.
\newblock \href {https://doi.org/10.13026/1n74-ne17} {{MIMIC-IV-Note}:
  Deidentified free-text clinical notes (version 2.2)}.
\newblock PhysioNet.

\bibitem[{Johnson et~al.(2023{\natexlab{c}})}]{johnson2023mimiciv}
Alistair E.~W. Johnson et~al. 2023{\natexlab{c}}.
\newblock {MIMIC-IV}, a freely accessible electronic health record dataset.
\newblock \emph{Scientific Data}, 10:1.

\bibitem[{Kwon et~al.(2023)}]{kwon2023vllm}
Woosuk Kwon et~al. 2023.
\newblock Efficient memory management for large language model serving with
  {PagedAttention}.
\newblock In \emph{SOSP}.

\bibitem[{Lin et~al.(2024)Lin, Tang, Tang, Yang, Chen, Wang
  et~al.}]{lin2024awq}
Ji~Lin, Jiaming Tang, Haotian Tang, Shang Yang, Wei-Ming Chen, Wei-Chen Wang,
  et~al. 2024.
\newblock {AWQ}: Activation-aware weight quantization for {LLM} compression and
  acceleration.
\newblock In \emph{Proceedings of Machine Learning and Systems (MLSys)}.
\newblock ArXiv:2306.00978.

\bibitem[{Matsuura et~al.(2026)Matsuura, Kumar, Kadam, and
  Weiss}]{hindsighttemporal2026}
Misaki Matsuura, Sayantan Kumar, Ojas Kadam, and Jeremy~C. Weiss. 2026.
\newblock Hindsight bias in clinical temporal reasoning: How future data
  exposure affects large language model judgment.
\newblock In \emph{Machine Learning for Health Symposium (ML4H)}.
\newblock ArXiv:2609.13454.

\bibitem[{McCoy et~al.(2019)McCoy, Pavlick, and Linzen}]{mccoy2019right}
R.~Thomas McCoy, Ellie Pavlick, and Tal Linzen. 2019.
\newblock Right for the wrong reasons: Diagnosing syntactic heuristics in
  natural language inference.
\newblock In \emph{Proceedings of ACL}.
\newblock ArXiv:1902.01007.

\bibitem[{Mu et~al.(2026)Mu, Huang, Gu, Qin, Zhang, and Zhang}]{ehrworld2026}
Linjie Mu, Zhongzhen Huang, Yannian Gu, Shengqian Qin, Shaoting Zhang, and
  Xiaofan Zhang. 2026.
\newblock {EHRWorld}: A patient-centric medical world model for long-horizon
  clinical trajectories.
\newblock \emph{arXiv preprint arXiv:2602.03569}.

\bibitem[{Nori et~al.(2025)Nori, Daswani, Kelly, Lundberg, Ribeiro, Wilson,
  Liu, Sounderajah, Carlson, Lungren, Gross, Hames, Suleyman, King, and
  Horvitz}]{nori2025sdbench}
Harsha Nori, Mayank Daswani, Christopher Kelly, Scott Lundberg, Marco~Tulio
  Ribeiro, Marc Wilson, Xiaoxuan Liu, Viknesh Sounderajah, Jonathan Carlson,
  Matthew~P. Lungren, Bay Gross, Peter Hames, Mustafa Suleyman, Dominic King,
  and Eric Horvitz. 2025.
\newblock Sequential diagnosis with language models.
\newblock \emph{arXiv preprint arXiv:2506.22405}.

\bibitem[{Pedregosa et~al.(2011)Pedregosa, Varoquaux, Gramfort, Michel,
  Thirion, Grisel et~al.}]{pedregosa2011sklearn}
Fabian Pedregosa, Ga{\"e}l Varoquaux, Alexandre Gramfort, Vincent Michel,
  Bertrand Thirion, Olivier Grisel, et~al. 2011.
\newblock Scikit-learn: Machine learning in {P}ython.
\newblock \emph{Journal of Machine Learning Research}, 12:2825--2830.

\bibitem[{Poliak et~al.(2018)Poliak, Naradowsky, Haldar, Rudinger, and
  Van~Durme}]{poliak2018hypothesis}
Adam Poliak, Jason Naradowsky, Aparajita Haldar, Rachel Rudinger, and Benjamin
  Van~Durme. 2018.
\newblock Hypothesis only baselines in natural language inference.
\newblock In \emph{Proceedings of the Seventh Joint Conference on Lexical and
  Computational Semantics (*SEM)}.
\newblock ArXiv:1805.01042.

\bibitem[{Qiu et~al.(2025)Qiu, Wu, Liu, Zheng, Liao, Wang, Yue, Fan, Zhen,
  Wang, Gu, Wang, Zhang, and Xie}]{qiu2025diaggym}
Pengcheng Qiu, Chaoyi Wu, Junwei Liu, Qiaoyu Zheng, Yusheng Liao, Haowen Wang,
  Yun Yue, Qianrui Fan, Shuai Zhen, Jian Wang, Jinjie Gu, Yanfeng Wang,
  Ya~Zhang, and Weidi Xie. 2025.
\newblock Evolving interactive diagnostic agents in a virtual clinical
  environment.
\newblock \emph{arXiv preprint arXiv:2510.24654}.

\bibitem[{{Qwen Team}(2024)}]{qwen2025}
{Qwen Team}. 2024.
\newblock Qwen2.5 technical report.
\newblock \emph{arXiv preprint arXiv:2412.15115}.

\bibitem[{Sainz et~al.(2023)Sainz, Campos, Garc{\'i}a-Ferrero, Etxaniz,
  Lopez~de Lacalle, and Agirre}]{sainz2023contamination}
Oscar Sainz, Jon~Ander Campos, Iker Garc{\'i}a-Ferrero, Julen Etxaniz, Oier
  Lopez~de Lacalle, and Eneko Agirre. 2023.
\newblock {NLP} evaluation in trouble: On the need to measure {LLM} data
  contamination for each benchmark.
\newblock In \emph{Findings of EMNLP}.
\newblock ArXiv:2310.18018.

\bibitem[{Schmidgall et~al.(2024)Schmidgall, Ziaei, Harris, Reis, Jopling, and
  Moor}]{schmidgall2024agentclinic}
Samuel Schmidgall, Rojin Ziaei, Carl Harris, Eduardo Reis, Jeffrey Jopling, and
  Michael Moor. 2024.
\newblock {AgentClinic}: a multimodal agent benchmark to evaluate {AI} in
  simulated clinical environments.
\newblock \emph{arXiv preprint arXiv:2405.07960}.

\bibitem[{Tang et~al.(2025)Tang, Yu, Su, Feng, Zhu, Wang, Liang, Zhang, Ding,
  and Chen}]{clindef2025}
Yuqi Tang, Jing Yu, Zichang Su, Kehua Feng, Zhihui Zhu, Libin Wang, Lei Liang,
  Qiang Zhang, Keyan Ding, and Huajun Chen. 2025.
\newblock {ClinDEF}: A dynamic evaluation framework for large language models
  in clinical reasoning.
\newblock \emph{arXiv preprint arXiv:2512.23440}.

\bibitem[{Yang et~al.(2026)Yang, Liu, Li, Li, Cao, Feng, and
  Li}]{rare2026audit}
Jian Yang, Peilin Liu, Xiaohan Li, Songda Li, Yu~Cao, Yuqing Feng, and Haomin
  Li. 2026.
\newblock Auditing evaluation artifacts and data leakage in large language
  model systems for rare disease diagnosis.
\newblock \emph{arXiv preprint arXiv:2610.33592}.

\end{thebibliography}

\clearpage
\onecolumn
\appendix

\section{Registered tests and decision rules}
\label{app:history}
Each failed test led to a design revision whose rationale was written down before the next
run. \textbf{D1} asked whether a fine-tuned simulator given the diagnosis (\splus{}) produces
results more predictable than real ones. \textbf{A} replaced it with a prompted gatekeeper.
A subsequent review identified four validity problems: \sminus{} itself carries hindsight,
sequential conditioning on real results leaves the simulator little room, 0/1 accuracy is
coarse, and the text judge is sensitive to format. \textbf{A2} therefore used independent
generation, the presentation-time control \szero{}, a numeric judge and \ptrue{}.
\textbf{A$'$} registered three tests jointly with Holm correction and $B=10{,}000$ bootstrap
resamples. \textbf{E1} was registered after the exploratory case-text analyses, to replicate
them on the validation split.

Table~\ref{tab:prereg-detail} gives each rule as registered, in the order the tests were run.
Every rule was recorded in a versioned project protocol before the corresponding run started.
The rules, their timing and the protocol's revision history are deposited at OSF
(\ifdefined\ARXIV\url{https://osf.io/m8ykh/overview?view_only=3e27f98df9224cf6aca1f2d3ecbbe38f}\else anonymized view-only link:
\url{https://osf.io/m8ykh/overview?view_only=3e27f98df9224cf6aca1f2d3ecbbe38f}\fi).

\begin{table}[h!]
\centering
\small
\begin{tabular}{@{}L{1.4cm}L{11.0cm}L{2.6cm}@{}}
\toprule
Test & Decision rule as registered & Outcome \\
\midrule
D1 & Stop if, after removing explicit leakage, the CI of \splus{}$-$real (clf top-1) includes 0 or the gain is $<2$ pp; both judges must agree. Implemented two-sided at first; corrected to the registered one-sided form before the decision. & stop (both judges) \\[2pt]
A & Pass if \splus{}$-$\sminus{} $\ge 2$ pp with CI lower bound $>0$ \emph{and} the \splus{}$-$real point estimate is $>0$ (clf top-1). Otherwise no third simulator is tested. & fail \\[2pt]
A2 & Pass if the CIs of \splus{}$-$real \emph{and} \splus{}$-$\szero{} both lie above 0 (num judge, \ptrue{}). Declared the last test of this hypothesis. & fail \\[2pt]
A$'$1--3 & Three one-sided tests, patient-level bootstrap ($B=10{,}000$), Holm at $\alpha=0.05$. H1: summary $-$ chief complaint, text-classifier \ptrue{}, filtered test cases. H3: \splus{}$-$real, num \ptrue{}, 32B gatekeeper with unit hints. H2: agent top-1, \splus{}$-$\szero{} simulator. & none rejected (smallest $p=0.320$) \\[2pt]
E1 & On the validation split, four-model mean top-1 gaps V1$-$V3 and V2$-$V3; one-sided paired bootstrap ($B=10{,}000$); replicated if the null is rejected for both under Holm at $\alpha=0.05$ and both gaps are $\ge$5 pp. & replicated \\
\bottomrule
\end{tabular}
\caption{Registered decision rules, in the order the tests were run. Analyses beyond these rules (DiagGym,
Phi-4, GK-72B, the presentation-time reconstruction, LLM diagnosis including Llama-3.1-8B and
the V3n format control, the stronger agent, the interactive agents and the cue analysis) were conducted afterwards and
are descriptive or exploratory.}
\label{tab:prereg-detail}
\end{table}

\section{Case construction and differences from DiagGym}
\label{app:diff}
DiagGym did not release its case-construction code, so we implemented the published
description. The main differences are:
\begin{enumerate}
  \item Physical-examination structuring and the ``diagnosis known before admission'' filter
  use a local Qwen2.5-7B-Instruct model instead of DeepSeek-V3.
  \item Diagnoses are grouped by AHRQ CCSR. Only emergency or urgent admissions are kept.
  Injury (INJ) and mental and behavioral (MBD) categories and transfers or post-procedure
  admissions are excluded.
  \item The fine-tuned simulator is Qwen2.5-3B with LoRA and diagnosis dropout rather than a
  fully fine-tuned 7B model. The public DiagGym 7B is evaluated separately.
  \item Laboratory items are kept at the single-analyte level (54.0 per case vs.\ 28.4).
\end{enumerate}

\section{Implementation details}
\label{app:impl}
Table~\ref{tab:impl} lists the settings of the simulators, judges, LLM diagnosis, agent and
statistics. Code and aggregate results will be released.

\begin{table}[!ht]
\centering
\small
\begin{tabular}{@{}L{3.2cm}L{12.2cm}@{}}
\toprule
Component & Setting \\
\midrule
FT-3B simulator & Qwen2.5-3B-Instruct + LoRA ($r=64$, $\alpha=128$, all linear layers); diagnosis dropout 0.5; lr $2\times10^{-4}$, cosine; 2 epochs; gradient accumulation 16; maximum length 6{,}144 tokens; loss on results only. \\
Simulated results & 5 samples per item, temperature 1.0, at most 512 new tokens; sequential (given real earlier results) or independent (profile only). \\
Numeric judge (num) & Multinomial logistic regression ($C=1$) on the 300 most frequent laboratory items: standardized value, abnormal flag, missing indicator; lenient number parsing. Values-only variant drops the flags. \\
Boosted judge & Histogram gradient boosting, at most 200 iterations, early stopping, same features as num. \\
Text judges & TF-IDF (1--2-grams, at most 30{,}000 features, min.\ document frequency 3, sublinear tf) + logistic regression ($C=1$). \\
LLM diagnosis & Temperature 0, at most 200 new tokens; one follow-up turn if no valid code is returned (afterwards at most 0.7\% of answers per model and view lacked a valid code); input truncated to 12{,}000 characters. Llama-3.1-8B-Instruct (bf16) was added as a third model family on the test split only. \\
V3n format control & V3 rewritten by Qwen2.5-14B-Instruct-AWQ with the V1 summary prompt (temperature 0, at most 800 new tokens); 1{,}635 test cases with a triage record, none empty. \\
GK-72B & Qwen2.5-72B-Instruct-AWQ, tensor-parallel over four GPUs, GPU memory fraction 0.72, at most 64 concurrent sequences; otherwise as GK-32B. \\
GK-14B (BF16) & Unquantized Qwen2.5-14B-Instruct, tensor-parallel over two GPUs; otherwise as GK-14B with independent generation (quantization check). \\
A$'$2 agent & Temperature 0; at most 5 panel orders; 3 invalid turns force a diagnosis. \\
Interactive agents & Six models (Table~\ref{tab:agentviews}); temperature 0, at most 20 new tokens per turn and 60 for the final answer; at most 8 requests among 7 actions; repeated or invalid requests count as invalid, and 3 invalid turns force a diagnosis; imaging report truncated to 1{,}500 characters; view truncated as for LLM diagnosis. \\
Statistics & Patient-level paired bootstrap, 95\% percentile CIs; $B=10{,}000$ for registered tests and LLM diagnosis, $B=2{,}000$ for other descriptive analyses. \\
Software and hardware & vLLM 0.7.3, PyTorch 2.5.1, Transformers 4.49, PEFT 0.14, scikit-learn; four RTX 4090 GPUs (24\,GB). \\
\bottomrule
\end{tabular}
\caption{Hyperparameters and settings.}
\label{tab:impl}
\end{table}

\section{Format and unit checks}
\label{app:units}
Numeric values were parseable (anywhere in the string) for 72.5\% of real and 90.8--94.6\% of
GK-14B results. Simulated results used the real unit in 41--43\% of cases, but most
differences were numerically equivalent (e.g.\ $10^9$/L vs.\ K/$\mu$L, mmol/L vs.\ mEq/L for
monovalent ions). Glucose was the main exception and was addressed by unit hints in A$'$3.

\section{Calibration of the A$'$1 classifiers (post hoc)}
\label{app:calib}
Table~\ref{tab:calib} compares the two classifiers of test A$'$1 on the 1{,}760 test cases
whose summary passed the leakage filter. The temperature $T$ was fitted by minimizing the
negative log-likelihood on the validation split. The registered verdict is unchanged.
The summary classifier was strongly under-confident (expected
calibration error 0.239 vs.\ 0.063 for the chief-complaint classifier, mean entropy 3.09 vs.\
2.73 nats). Under the negative log-likelihood, a proper scoring rule, the summary was more
informative even without recalibration (gain 0.284 nats [0.231, 0.339]). After temperature
scaling fitted on the validation split, the \ptrue{} difference was $+0.131$ [$+0.117$,
$+0.145$]. The summary gain was broad (higher top-1 in 37 of 50 categories, median
$+10.1$ pp).

\begin{table}[h!]
\centering
\small
\begin{tabular}{@{}lccccccc@{}}
\toprule
Classifier & $T$ & top-1 & mean \ptrue{} & mean max-prob & entropy & ECE & NLL \\
\midrule
Summary, raw              & 1.000 & 0.491 & 0.192 & 0.252 & 3.089 & 0.239 & 2.122 \\
Summary, scaled           & 0.561 & 0.491 & 0.352 & 0.510 & 1.874 & 0.057 & 1.776 \\
Chief complaint, raw      & 1.000 & 0.357 & 0.190 & 0.294 & 2.734 & 0.063 & 2.406 \\
Chief complaint, scaled   & 0.856 & 0.357 & 0.221 & 0.355 & 2.430 & 0.038 & 2.375 \\
\bottomrule
\end{tabular}
\caption{The two A$'$1 classifiers before and after temperature scaling.}
\label{tab:calib}
\end{table}

\section{Sensitivity to explicit admission examinations}
\label{app:sens}
Restricting evaluation to cases whose physical-examination section explicitly marks an
admission examination (keyword label, 299 of 460 generation cases, 1{,}716 of 2{,}748
validation cases, 1{,}098 of 1{,}760 filtered test cases) left every conclusion unchanged. This check was run on the simulators available at the time
(GK-14B and GK-32B for independent generation): retention of their \splus{} results was
0.06--0.19 with the linear judge (values and flags) and 0.40--0.60 with the boosted judge. The contrast \splus{}$-$\szero{} in \ptrue{} was $+0.010$ to $+0.027$ (the
linear-judge GK-14B interval touched 0). History versus chief complaint top-1 was $+20.2$ pp
[$+17.6$, $+22.8$], and the summary versus chief complaint NLL gain was 0.317 nats [0.246, 0.390].

\section{Which words carry the signal}
\label{app:cues}
\begin{figure}[t]
\centering
\includegraphics[width=0.55\textwidth]{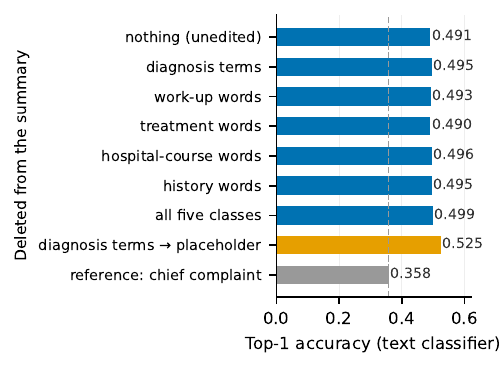}
\caption{Top-1 accuracy of the summary text classifier after deleting classes of words from the
training and test summaries (1{,}760 filtered test cases). Replacing diagnosis terms with a
placeholder (orange) increases accuracy, an artifact discussed in Appendix~\ref{app:pitfalls}. Dashed line: chief
complaint alone.}
\label{fig:cues}
\end{figure}

Table~\ref{tab:cues} lists, for ten categories, the terms with the largest weights in the
summary text classifier (only terms occurring in at least 50 training summaries, aggregate
vocabulary, no patient text). They are mostly symptom descriptions and synonyms or
abbreviations that a list of diagnosis names does not cover. Figure~\ref{fig:cues} shows the
text-classifier accuracy after deleting each class of words.

\begin{table}[h!]
\centering
\small
\begin{tabular}{@{}L{4.6cm}L{10.8cm}@{}}
\toprule
Category (CCSR) & Largest-weight terms \\
\midrule
Acute pulmonary embolism & pulmonary, pleuritic, dyspnea, \textbf{clots}, chest, pulmonary embolism, \textbf{pe}, embolism \\
Gastrointestinal hemorrhage & \textbf{melena}, stools, \textbf{black}, bleeding, stool, \textbf{dark}, blood, gi \\
Intestinal obstruction and ileus & \textbf{flatus}, bowel, abdominal, abdominal pain, small bowel, obstruction, bowel movement \\
Chronic kidney disease & \textbf{esrd}, kidney, blood pressure, renal, stage, kidney disease, \textbf{dialysis} \\
Cardiac dysrhythmias & \textbf{palpitations}, atrial, fibrillation, tachycardia, \textbf{atrial fibrillation}, rate \\
Heart failure & dyspnea, heart failure, \textbf{edema}, \textbf{orthopnea}, heart, failure, of breath \\
Biliary tract disease & \textbf{ruq}, \textbf{jaundice}, cholecystitis, cholangitis, upper quadrant, right upper, \textbf{ercp} \\
Thromboembolism (phlebitis, DVT) & swelling, \textbf{dvt}, leg, extremity, \textbf{calf}, vein, lower extremity \\
Cerebral infarction & stroke, speech, weakness, \textbf{sided weakness}, \textbf{aphasia}, difficulty, hand \\
COPD and bronchiectasis & \textbf{copd}, cough, dyspnea, sputum, history copd, \textbf{wheezing}, productive \\
\bottomrule
\end{tabular}
\caption{Largest-weight terms of the summary classifier for ten categories. Bold: synonyms,
abbreviations or characteristic symptom descriptions not caught by a diagnosis-name filter.}
\label{tab:cues}
\end{table}

\paragraph{An invented illustration.} The following pair is \emph{invented} for illustration and
is not derived from any patient record. A presentation-time view might read: ``Female, 68.
Chief complaint: weakness. Triage: HR 112, BP 98/60, pain 0. Past medical history: hypertension,
osteoarthritis.'' A summary written from the discharge record of the same (fictional) admission
might read: ``68-year-old woman with two days of black, tarry stools and lightheadedness on
standing, taking daily ibuprofen for knee pain.'' The summary does not name the bleeding, yet
\emph{black, tarry stools} and the NSAID use point to gastrointestinal hemorrhage. A filter that
looks for the diagnosis name would therefore accept it. Both details could have been elicited at the bedside.
From the text alone one cannot tell whether they were reported at arrival or selected because
the answer was known when the summary was written. This is why the V1$-$V3 contrast is an upper
bound on hindsight rather than an estimate of it, and why a name filter, which detects neither
possibility, cannot address it. The example thus illustrates a lexical shortcut. It does not by itself demonstrate
hindsight.

\section{Additional results for the case-text channel}
\label{app:casetext}
\paragraph{History versus chief complaint by category.} History was more predictive in 43 of 50 categories (median gain $+17.8$ pp). The largest gains
were for secondary malignancies ($+63$ pp), other complications of pregnancy ($+62$),
asthma ($+56$), inflammatory bowel disease ($+50$), pancreatic ($+45$) and biliary tract
disease ($+42$). The history was \emph{less} predictive than the chief complaint for
symptom-defined categories such as nausea and vomiting ($-37$), nonspecific chest pain
($-14$) and hypotension ($-9$), where the complaint is itself the label.

\begin{figure*}[t]
\centering
\includegraphics[width=0.82\textwidth]{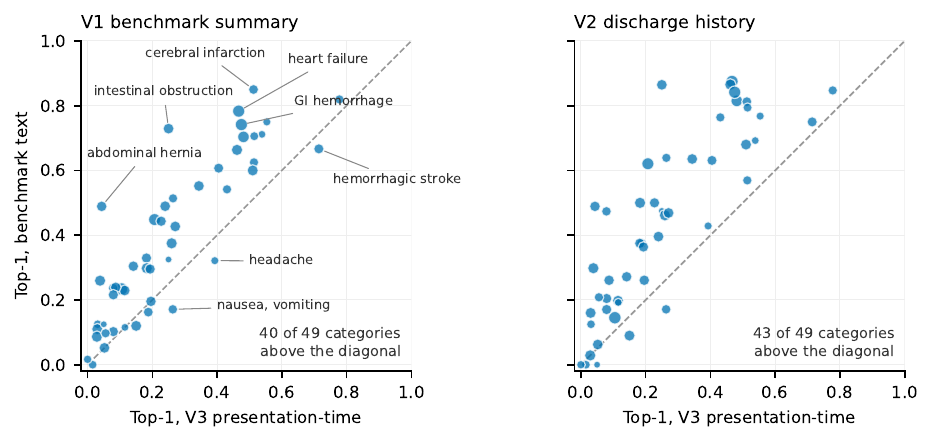}
\caption{Per-category top-1 accuracy of zero-shot diagnosis from the benchmark text (left: V1
summary, right: V2 discharge history) against the presentation-time view V3, mean of the four
LLMs, for the 49 CCSR categories with at least ten test cases (marker area proportional to the
number of cases). Points above the dashed diagonal are categories where the benchmark text is
easier. Exploratory, pilot test cases. Categories have 10--31 cases, so individual estimates are imprecise.}
\label{fig:category}
\end{figure*}

\paragraph{Robustness and breadth.} With a differently worded prompt (no persona,
candidates listed first), all accuracies fell and the long discharge history was affected
most, but the summary still raised accuracy by 10.0 pp [7.2, 12.8]. The magnitude is
prompt-sensitive while the direction is not. The summary beat the presentation-time view in
82\% of the 49 categories with at least ten cases (median +11.5 pp, Figure~\ref{fig:category}). Gains were largest for
diagnoses typically confirmed by imaging or a procedure, such as intestinal obstruction (+48 pp),
abdominal hernia (+45), cerebral infarction (+34), gastrointestinal hemorrhage (+27) and
spondylopathies (+25), as well as for heart failure (+32). They were absent for symptom-defined
categories (nausea and vomiting $-9$, headache $-7$, nonspecific chest pain $-3$) and for
hemorrhagic stroke ($-5$).

\paragraph{Interactive agents.}\label{app:agentviews} Table~\ref{tab:agentviews} gives the
results of the six interactive agents of \S\ref{sec:v3} (temperature 0, at most eight requests
among CBC, CHEM, LFT, LACTATE, UA, IMAGING and EXAM; results are replayed from the record, and a
request without a recorded result returns ``not done''; three invalid turns force a diagnosis;
the final turn asks for the five most likely of the 50 groups; prompts in
Appendix~\ref{app:prompts}). The agents made 4.8--6.7 requests per case and at most 0.1\% of
final answers per agent and view could not be parsed. This analysis is exploratory and uses the
same pilot test cases as Table~\ref{tab:llmdx}. Before running it, we fixed an internal decision
rule for extending the study: an average V1$-$V3 gap of at least 5 pp \emph{and} at least one
pair of agents whose order reversed between V1 and V3 by at least 3 pp in both views. The first
condition was met and the second was not (the only reversal, Phi-4 and Qwen2.5-32B, was 0.7 and
2.6 pp), so we report the score inflation together with the stable ranking.

\begin{table}[h!]
\centering
\small
\begin{tabular}{@{}lrrrrrc@{}}
\toprule
Agent & V1 & V2 & V3 & V1$-$V3 & V2$-$V3 & requests per case (V1/V2/V3) \\
\midrule
Qwen2.5-7B        & 33.0 & 38.7 & 24.6 & +8.4 & +14.1 & 6.1 / 6.1 / 6.0 \\
Llama-3.1-8B      & 29.4 & 36.1 & 22.5 & +6.9 & +13.6 & 6.1 / 5.7 / 6.0 \\
Qwen2.5-14B-AWQ   & 38.5 & 44.3 & 29.9 & +8.6 & +14.4 & 5.8 / 4.8 / 5.4 \\
Phi-4             & 40.5 & 45.0 & 35.6 & +4.9 & +9.4 & 6.7 / 6.3 / 6.7 \\
Qwen2.5-32B-AWQ   & 41.2 & 46.5 & 33.0 & +8.2 & +13.5 & 6.0 / 5.7 / 6.0 \\
Qwen2.5-72B-AWQ   & 44.6 & 48.7 & 38.7 & +5.9 & +10.0 & 6.2 / 6.3 / 6.2 \\
\midrule
Mean (six)        & 37.9 & 43.2 & 30.7 & +7.2 & +12.5 & \\
\bottomrule
\end{tabular}
\caption{Top-1 accuracy (\%) of six interactive agents reading V1 (benchmark summary), V2
(discharge history) or V3 (presentation-time reconstruction) of the 1{,}053 test cases.
Per-agent differences are computed from the rounded accuracies; the means and their 95\%
patient-level bootstrap CIs (V1$-$V3 [5.6, 8.8], V2$-$V3 [10.8, 14.2]) from unrounded values.
Ranking by V1 and V2: 72B $>$ 32B $>$ Phi-4 $>$ 14B $>$ 7B $>$ Llama; by V3, Phi-4 and 32B swap
(Kendall $\tau=0.87$).}
\label{tab:agentviews}
\end{table}

\begin{figure*}[t]
\centering
\includegraphics[width=0.85\textwidth]{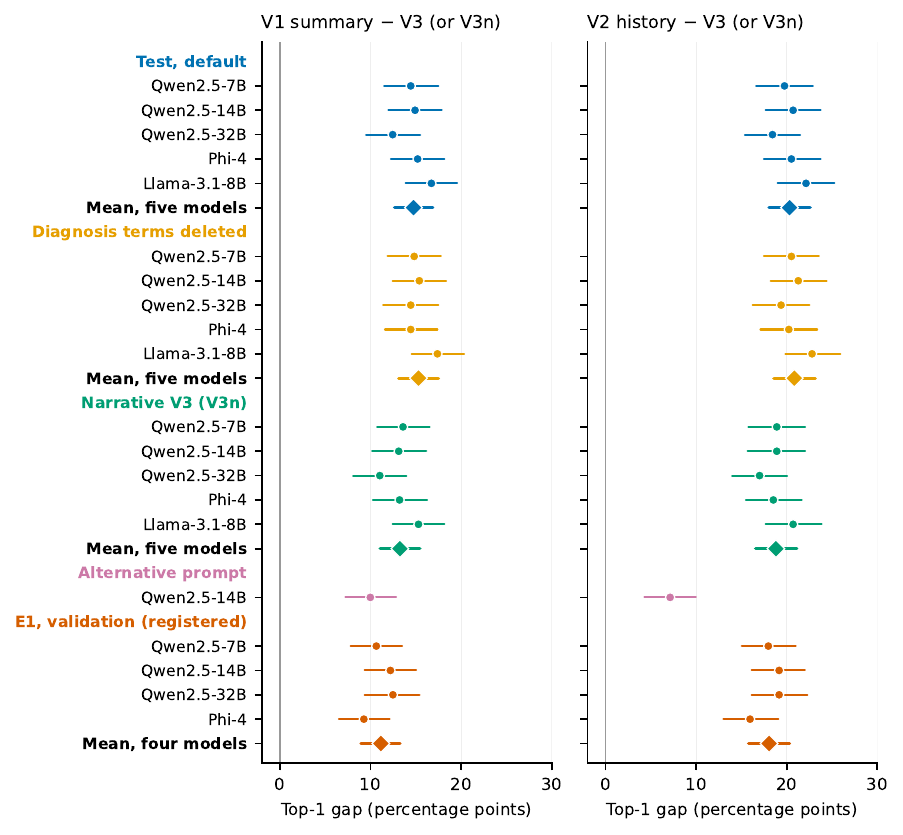}
\caption{Top-1 accuracy gap of zero-shot LLM diagnosis between the benchmark text (left: V1
summary, right: V2 discharge history) and the presentation-time view V3, per model and
condition, with 95\% patient-level bootstrap CIs. For V3n, the presentation-time view is V3
rewritten as a narrative by the model and instruction that produced V1. E1 is the registered
replication on the validation split. All other rows are exploratory analyses of the 1{,}053 test
cases. Means are taken over the models listed in each block.}
\label{fig:forest}
\end{figure*}

\section{Additional simulator results}
\label{app:simextra}
\paragraph{Quantized versus unquantized gatekeeper.} To test whether dilution stems from AWQ
quantization, we reran GK-14B with the unquantized Qwen2.5-14B-Instruct (BF16, tensor-parallel
over two GPUs) under exactly the same settings (independent generation, \splus{}, \sminus{} and
\szero{}, 460 cases, five samples per item) and evaluated both versions with the same judges.
Retention of \splus{} results was 0.19 [0.09, 0.30] for BF16 and 0.16 [0.06, 0.27] for AWQ with
the values-only judge, and 0.47 [0.38, 0.57] and 0.43 [0.34, 0.53] with the boosted judge.
\splus{}$-$\szero{} was +0.023 [+0.011, +0.036] and +0.019 [+0.011, +0.028]. The diagnosis line
again had a negligible effect (\splus{}$-$\sminus{} +0.002 in \ptrue{}). The BF16 results were somewhat
shorter and contained interpretive words less often (2.4\% vs.\ 3.6\% of results), but the
dilution was essentially unchanged.

\paragraph{No sign of memorization.} DiagGym was also run on 225 of the 750 DiagBench cases its
authors held out from training, restricted to our 50 categories and to patients outside our
judges' training data. It retained as much signal on the 225 held-out
DiagBench cases it was not trained on (\splus{} 70\% [55, 88]) as on our cases, which may
overlap its training data (72\% [60, 86]). The two case sets differ, so this is a coarse
check.

\begin{figure*}[t]
\centering
\includegraphics[width=0.85\textwidth]{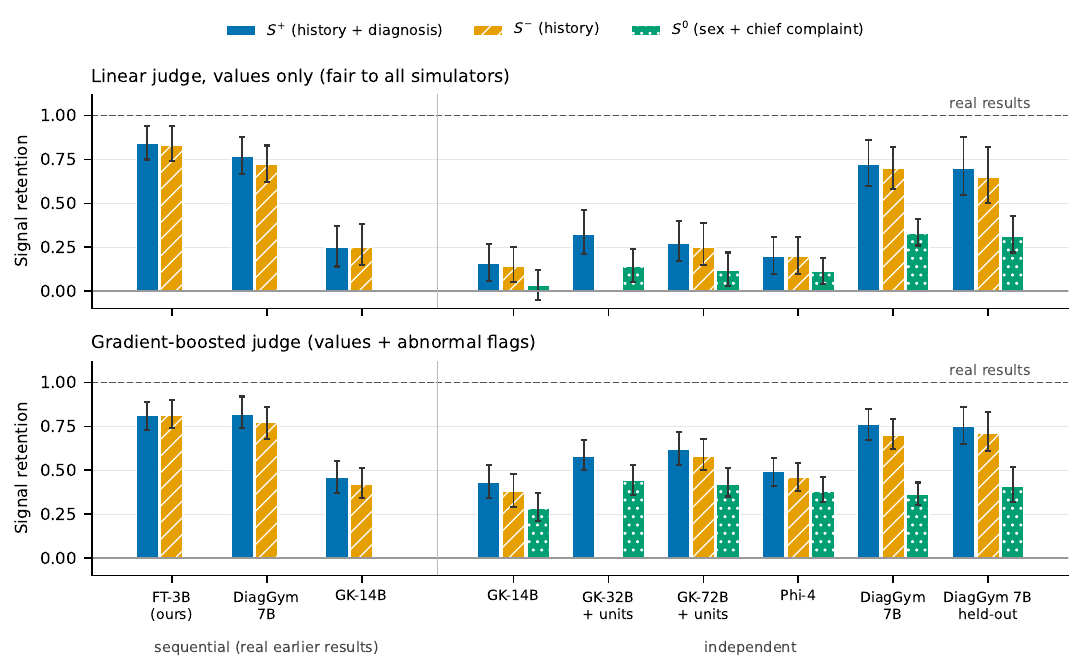}
\caption{Signal retention of simulated laboratory results, $(\bar p_{\text{sim}}-c)/(\bar p_{\text{real}}-c)$
with $\bar p$ the mean probability a real-data judge assigns to the true category and
$c=1/50$, so that 1 means as informative as the real results. 460 cases (held-out: 225 DiagBench
cases DiagGym was not trained on), five samples per item, 95\% patient-level bootstrap CIs.
Sequential runs condition each result on the real earlier results and are not comparable
with independent runs. Top: logistic regression on values and missingness only, fair to
simulators that never write abnormal flags. Bottom: gradient-boosted trees on values,
flags and missingness. All judges are trained on real data only.}
\label{fig:retention}
\end{figure*}

\begin{table}[t]
\centering

\small
\begin{tabular}{@{}lccc@{}}
\toprule
 & \splus{} & \sminus{} & \szero{} \\
\midrule
GK-14B          & 0.454 & 0.438 & 0.359 \\
GK-32B + units  & 0.612 & --    & 0.394 \\
GK-72B + units  & 0.609 & 0.582 & 0.534 \\
Phi-4 + units   & 0.618 & 0.598 & 0.481 \\
\bottomrule
\end{tabular}
\caption{Typicalization: share of real results that are atypically normal for the diagnosis
that a simulator flips to the typical abnormal state (independent generation). Not
computable for DiagGym, which never writes abnormal flags.}
\label{tab:typical}
\end{table}

\paragraph{Full agent results.} The agent reached top-1 of
20.7\% with real results and 19.3\% with either gatekeeper (top-5: 40.1\%, 40.4\%,
40.9\%). \splus{}$-$real was $-1.39$ pp [$-3.33$, $+0.42$]. The agent ordered about three
panels per case in all conditions. A descriptive follow-up explains why: with no laboratory
results at all, the same agent reached 19.6\% (real $-$ none $+0.8$ pp [$-1.5$, $+3.2$]), indicating that
it made little use of laboratory results. A stronger agent (Qwen2.5-32B-AWQ, also as
gatekeeper) benefited slightly from them (real $-$ none $+1.9$ pp [$0.0$, $+3.9$]), and its accuracy still did not depend on the
simulator (\splus{}$-$\szero{} $+0.6$ pp [$-1.0$, $+2.1$], \splus{}$-$real $0.0$). In this
setting, laboratory results changed agent accuracy by at most about two points, an order of magnitude
less than the case text does (\S\ref{sec:v3}). Because these agents make little use of laboratory
results, A$'$2 cannot show whether simulator leakage would matter for an agent that does. Our
conclusions about simulators rest on the real-data judges (\S\ref{sec:prereg} and below), not on
agent scores.

\paragraph{Laboratory-defined subset.} In a
post hoc subset of the eight categories usually established by the five panels (diabetes with
complication, fluid and electrolyte disorders, acute renal failure, chronic kidney disease, other
liver disease, hepatitis, urinary tract infection and septicemia, 175 cases in total, with the list fixed
before per-category agent results were inspected), the 32B agent did use the laboratory results
(real $-$ none +4.0 pp [0.6, 8.0]), and its accuracy depended on the simulator: \splus{}$-$\szero{}
was +4.0 pp [1.1, 7.4] in top-1 and +9.1 pp [1.7, 16.6] in top-5, and \splus{}$-$real was +1.7 pp
[$-1.1$, 4.6] in top-1 and +8.0 pp [2.3, 13.7] in top-5. The 14B agent did not use the results in
this subset either (real $-$ none +1.7 pp [$-2.3$, 6.3]), and in its registered runs with the 14B
gatekeeper the simulator made no difference there (\splus{}$-$\szero{} +0.6 pp [$-1.7$, 2.9],
\splus{}$-$real $-1.1$ pp [$-4.6$, 2.3]). These contrasts are exploratory, not
corrected for multiple comparisons, and confound the diagnosis line with the retrospective
profile, because the agent runs had no \sminus{} arm. They nevertheless suggest that, where an
agent relies on laboratory results, a gatekeeper holding the hindsight case file can raise its
score above real replay, although for a judge trained on real data the same simulated results
carry less signal than real ones in these categories (\ptrue{} 0.110 vs.\ 0.264, values-only
judge). The agent reads the full text of each simulated result, whereas the judges see only parsed
values, but accompanying wording does not explain the advantage: in the 32B gatekeeper's
results for these categories, interpretive words such as \emph{elevated} or \emph{consistent with}
occurred in 0.4\% of results under both \splus{} and \szero{}, and diagnosis terms in 0.10\% and
0.06\%. A more plausible source is typicalization. Given the retrospective profile and the diagnosis, the 32B gatekeeper
turned 61\% of atypically normal real results into the typical abnormal state, against 39\% with
\szero{} (Table~\ref{tab:typical}). This can supply the few decisive values an agent looks for,
while the remaining values carry less signal for a judge that uses all 300 items. We state this
as a hypothesis.

\section{Pitfalls for leakage audits}
\label{app:pitfalls}
Both arose in our own analyses. First,
masking diagnosis terms with a placeholder, as is common practice~\cite{hager2024mimiccdm},
gave text classifiers a new cue: masked summaries became \emph{more} predictable than
unmasked ones (top-1 0.525 vs.\ 0.491), and masked histories likewise (0.574 vs.\ 0.559),
whereas deleting the terms changed little (0.495 and 0.558). A likely mechanism is that the
number and position of placeholders encode how often and where the diagnosis was named, which
differs between categories. For the LLMs the choice
hardly mattered (summary gap 14.4 vs.\ 14.8 pp for Qwen2.5-7B and 15.3 vs.\ 15.4 for
-14B with placeholder vs.\ deletion), but judges that treat the placeholder as a token are
affected. Second, our registered test of the summaries compared the mean probability
of the true class across two \emph{different} classifiers and found no difference ($+0.002$), although
accuracy rose by 13.4 pp, because the summary classifier was less confident (ECE 0.239 vs.\
0.063), and a proper scoring rule (negative log-likelihood) or temperature scaling recovers the
effect (Appendix~\ref{app:calib}). Comparisons across classifiers should use accuracy or proper
scores, not raw probabilities.

\section{Recommendations}
\label{app:recs}
Table~\ref{tab:checklist} lists practical recommendations for building and auditing clinical agent
benchmarks. Each recommendation is linked to the result that motivates it. Recommendations 1--3
concern the agent-facing case text, 4--6 the simulator, and 7 the design of leakage audits. They
are derived from one data source, 50 common emergency categories, laboratory panels and the open
models we could run locally, and should be revisited for other settings (Limitations).
\begin{table*}[t]
\centering
\small
\begin{tabular}{@{}L{0.6cm}L{7.1cm}L{7.1cm}@{}}
\toprule
& Recommendation & Evidence in this paper \\
\midrule
1 & Build agent-facing cases from information documented at presentation (e.g.\ triage records), or report a presentation-time baseline next to every score. & Benchmark-style text made zero-shot diagnosis 11--20 pp easier than a presentation-time reconstruction, on test and held-out patients (\S\ref{sec:v3}). \\
2 & Do not rely on diagnosis-name filters; test filters against implicit cues. & Deleting diagnosis terms, alone or with four other word classes, left the gap intact (Table~\ref{tab:llmdx}, Figure~\ref{fig:cues}). \\
3 & When auditing with text-based judges, mask by deletion rather than by a placeholder token. & A placeholder raised summary top-1 from 0.491 to 0.525 (Appendix~\ref{app:pitfalls}). \\
4 & Report the signal retention of simulated results against real results, using a judge trained on real data only. & Prompted gatekeepers retained 16--62\% of the real signal and trained world models 70--84\% (\S\ref{sec:retention}). \\
5 & Condition simulators on presentation-time profiles where possible, and report the gap to retrospective profiles. & Retrospective profiles increased predictability by 0.014--0.059 in \ptrue{}, whereas the diagnosis line had a negligible effect (\S\ref{sec:retention}). \\
6 & Before attributing score differences to simulators, verify that the agent uses the simulated results. & Overall, agent accuracy differed by at most about two points with and without laboratory results; in a laboratory-defined subset, a stronger agent scored higher with the hindsight gatekeeper than with real results (\S\ref{sec:prereg}). \\
7 & Register audits before running them, and compare classifiers by accuracy or proper scoring rules rather than raw probabilities. & A$'$1 compared raw probabilities across classifiers and missed a 13.4-pp accuracy gain (\S\ref{sec:casetext}, Appendix~\ref{app:calib}). \\
\bottomrule
\end{tabular}
\caption{Recommendations for building and auditing clinical agent benchmarks.}
\label{tab:checklist}
\end{table*}

\section{Compute and data governance}
\label{app:compute}
Every model in this study runs on a local server accessible only to credentialed project members
(vLLM~\cite{kwon2023vllm}). No record text is sent to any external service. This is our
data-governance choice for MIMIC text, and it excludes proprietary API models such as the
o4-mini gatekeeper of SDBench. We therefore report how each result changes with model size
and family (3B to 72B parameters from the Qwen2.5, Phi-4 and Llama-3.1 families) rather than with a frontier API model.

All experiments ran on one server with four NVIDIA RTX 4090 GPUs (24\,GB each). Jobs that
completed successfully used about 155 GPU-hours in total (85 hours of wall-clock time, excluding failed and
restarted runs). The largest items were the six interactive agents reading V1--V3 (62 GPU-hours, of
which 43 for the 72B agent), the 72B gatekeeper (31), the other
simulators and gatekeepers (14B, 32B, Phi-4 and DiagGym: 23), zero-shot LLM diagnosis over all
models, views and variants including V3n (14), fine-tuning FT-3B (10), DiagGym-style summaries and
preprocessing with local LLMs (7), the A$'$2 agents (6) and the LLM judge (1). Judges, text classifiers
and statistics ran on CPU. Data were used under the PhysioNet credentialed data use agreement. No
record-level text left the server, and all reported numbers are aggregates.

\section{Prompts}
\label{app:prompts}
All prompts are reproduced from the code. Placeholders in braces are filled per case, and
\texttt{\{candidates\}} lists the 50 CCSR codes with their descriptions. The V3n format control
uses the V1 summary prompt below, with V3 as the input record.
\needspace{8\baselineskip}
\subsection{Simulator: gatekeeper (system and user message; \splus{} adds the final diagnosis to the case file)}
\begin{lstlisting}
You are the information gatekeeper of a clinical diagnosis benchmark. You hold the patient's case file. When an examination is requested, report its result as it would appear in the hospital record. If the result is not in the case file, write a realistic, clinically plausible result that is consistent with the case. Report only the result itself, in the same style as the results reported so far (for laboratory tests: value and unit, followed by ' [abnormal]' if outside the reference range; for imaging: a short report). Do not explain.

Case file:
{case}

Results reported so far:
{past}

Requested examination: {name}
Result:
\end{lstlisting}

\needspace{8\baselineskip}
\subsection{Pre-hospital summary: case-summary part of DiagGym's prompt~3 (verbatim)}
\begin{lstlisting}
As an experienced physician, you will receive a patient Electronic Health Record focused on diagnosis. Your task is to:
- Summarize only the key clinical information available prior to hospital presentation (i.e., the patient's state before arrival at the hospital, including symptoms, history, and relevant background). Do not include any information from the hospital course, ancillary tests, laboratory or imaging results, treatments, procedures, or discharge summaries.
Ensure all summaries are concise, accurate, and based solely on the information provided. Do not reference images, tables, or other visual data, as these are not accessible.
Format to follow:
### Case Summary
Provide a detailed medical history of the patient prior to hospital arrival, including chief complaint, history of present illness, past medical history, family history, and any other relevant information for initial diagnosis. Do not include any findings, investigations, or events that occurred after hospital arrival. Do not include any ancillary tests or pertinent results here. Do not mention or imply any diagnostic conclusions. Do not include any language that attributes symptoms to a specific diagnosis, even if this is present in the EHR.
- Patient Information:
- Chief Complaint: If none, write "None."
- History of Present Illness: If none, write "None." This needs to be done very carefully and should only include information from before the visit.
- Past Medical History: If none, write "None."
- Personal History: If none, write "None."
- Family History: If none, write "None."
- Allergy History: If none, write "None."
...(other necessary information before hospital arrival)

Output only the "### Case Summary" section.

The following is the Electronic Health Record (EHR) of the patient:
{ehr_text}
\end{lstlisting}

\needspace{8\baselineskip}
\subsection{Leakage filter: criterion~1 of DiagGym's prompt~4 (verbatim)}
\begin{lstlisting}
Please evaluate whether the case summary provided below is qualified diagnostic task data according to the following standards.
### Evaluation Criteria
1. Information Leakage
Does the "Case Summary" section directly contain the name or diagnostic results of the "Diagnose Results"? If diagnostic results appear directly (That is, without any reasoning), it is judged as "Unqualified".
Output Format:
Information Leakage: xxx
The following is specific information:
Case Summary: {case_summary}
Diagnose Results: {diagnosis_results}
\end{lstlisting}

\needspace{8\baselineskip}
\subsection{Agent (system message and final turn)}
\begin{lstlisting}
You are an emergency physician. A patient has just arrived. You can order laboratory panels, one per turn, before making a diagnosis. Available panels: CBC (complete blood count), CHEM (basic chemistry: electrolytes, BUN, creatinine, glucose), LFT (liver function tests), LACTATE (lactate), UA (urinalysis). Each turn reply with exactly one line: 'ORDER: <panel code>' or 'DIAGNOSE' when you are ready. At most 5 panels.

Now give the diagnosis. Candidate diagnosis groups:
{candidates}

Answer with the codes of the 5 most likely groups, most likely first, as a JSON list. Output only the list.
\end{lstlisting}

\needspace{8\baselineskip}
\subsection{Interactive agents reading V1--V3 (system message, first user turn; the final turn is as above)}
\begin{lstlisting}
You are an emergency physician. You will read the information available about a patient. You can then request examinations, one per turn, before making a diagnosis. Available: CBC (complete blood count), CHEM (basic chemistry), LFT (liver function tests), LACTATE, UA (urinalysis), IMAGING (the first imaging study performed and its report), EXAM (physical examination on admission). Each turn reply with exactly one line: 'ORDER: <code>' or 'DIAGNOSE' when you are ready. At most 8 requests.

Case information:
{view}
\end{lstlisting}

\needspace{8\baselineskip}
\subsection{LLM diagnosis, default prompt (system and user message)}
\begin{lstlisting}
You are an experienced emergency physician.

A patient presents to the emergency department. Case information:

{case}

Candidate diagnosis groups:
{candidates}

Which group is most likely the principal diagnosis of this hospital admission? Answer with the codes of
the 5 most likely groups, most likely first, as a JSON list, for example ["ABC001", "ABC002", "ABC003",
"ABC004", "ABC005"]. Output only the list.
\end{lstlisting}

\needspace{8\baselineskip}
\subsection{LLM diagnosis, alternative prompt}
\begin{lstlisting}
Candidate diagnosis groups (code: description):
{candidates}

Clinical information available about a hospital admission:
{case}

Rank the five candidate groups that best match the principal diagnosis, best first.
Reply with a JSON list of five codes only.
\end{lstlisting}

\end{document}